\documentclass{article}

\PassOptionsToPackage{numbers}{natbib}
\usepackage[preprint]{neurips_2026}

\usepackage[utf8]{inputenc}
\usepackage[T1]{fontenc}
\usepackage{hyperref}
\usepackage{url}
\usepackage{booktabs}
\usepackage{rotating}
\usepackage{amsfonts}
\usepackage{amsmath}
\usepackage{nicefrac}
\usepackage{microtype}
\usepackage{xcolor}
\usepackage{graphicx}
\usepackage{multirow}
\usepackage{subcaption}
\usepackage{enumitem}
\usepackage{pifont}
\usepackage{tikz}
\usetikzlibrary{positioning,arrows.meta,calc}
\usepackage{float}
\usepackage{etoc}
\newcommand{\myparagraph}[1]{\vspace{0pt}\noindent{\bf #1}}
\newcommand{\projectname}{\textsc{NeuroQA}}

\definecolor{imggrounded}{HTML}{7030A0}
\definecolor{imginformed}{HTML}{C04F15}
\newcommand{\ig}[1][image-grounded]{\textcolor{imggrounded}{#1}}
\newcommand{\ii}[1][image-informed]{\textcolor{imginformed}{#1}}

\title{NeuroQA: A Large-Scale Image-Grounded Benchmark for 3D Brain MRI Understanding}

\author{
    \textbf{Mohammad H. Abbasi}$^{1,*}$,
    \textbf{Favour Nerrise}$^{1,*}$,
    \textbf{Shaurnav Ghosh}$^1$,
    \textbf{Ridvan Yesiloglu}$^1$,\\
    \textbf{Yuncong Mao}$^1$,
    \textbf{Bailey Trang}$^1$,
    \textbf{Mohammad Asadi}$^1$,
    \textbf{Merryn Daniel}$^1$,\\
    \textbf{Gustavo Chau Loo Kung}$^1$,
    \textbf{Ken Chang}$^1$,
    \textbf{Pavan Pinkesh Shah}$^1$,
    \textbf{Adam Turnbull}$^1$,\\
    \textbf{Kyan Younes}$^1$,
    \textbf{Seena Dehkharghani}$^1$,
    \textbf{Ehsan Adeli}$^{1,\dagger}$\\
    $^1$Stanford University, Stanford, CA, USA\\
    $^*$Equal contribution\\
    $^\dagger$Corresponding author\\
\texttt{\{mabbasi, fnerrise, eadeli\}@stanford.edu}\\
\vspace{0.5em}
\small
\url{https://neuroqa.stanford.edu}
}

\begin{document}

\maketitle
\etocdepthtag.toc{mainmatter}

\begin{abstract}
We present \projectname{}, a large-scale benchmark for visual question answering in 3D brain magnetic resonance imaging (MRI), with 56,953 QA pairs from 12,977 subjects across 12 datasets. It spans ages 5--104 and five clinical domains: Alzheimer's, Parkinson's, tumors, white matter disease, and neurodevelopment. Unlike prior medical Visual Question Answering (VQA) efforts that operate on 2D slices or rely on narrow diagnostic labels, \projectname{} pairs every item with a full 3D volume. It evaluates 11 clinically grounded reasoning skills through three answer formats: Yes/No, multiple-choice, and open-ended. Of the 203 templates, 131 are \emph{\ig} (answerable from a 3-plane viewer) and 72 are \emph{\ii} (ground truth derived from quantitative volumetry or clinical instruments). To remove text-only shortcuts, we apply answer-distribution refinement, reducing closed-format text-only accuracy from $>$80\% to 44.6\%; image necessity itself is assessed separately through an image-grounding protocol released with the benchmark. The benchmark is built via a 38-rule deterministic QA pipeline, validated through two rounds of expert review and seven rounds of automated verification. Every QA pair is verified against its underlying FreeSurfer measurement, structured metadata, or structured radiology report fields, with zero same-subject contradictions across templates. 
To bridge the gap between AI performance and clinical expertise, we conduct a clinician evaluation in which two clinicians independently assess 100 test items on the frozen benchmark using a three-plane brain viewer. On closed-format (Yes/No + multiple-choice) test-public items, the best zero-shot vision-language model and a supervised 3D CNN baseline reach 47.5\% and 43.7\% accuracy respectively, both below the 49.4\% text-only majority-template floor. \projectname{} adopts a two-tier release architecture with public QA pairs for open-access datasets and reproducible generation scripts for datasets restricted by data use agreements (DUAs). The release also includes subject-level data splits, a held-out private test set, and an online leaderboard for standardized model comparison.
\end{abstract}

\section{Introduction}
Brain MRI is a primary modality for diagnosing neurological disease across the lifespan, from Alzheimer's disease \cite{jack1999prediction, cole2019brain} and white-matter pathology \cite{habes2016white} to longitudinal progression \cite{yuksel2018longitudinal} and hemispheric-asymmetry analyses \cite{kong2018mapping}. Unlike 2D chest X-ray or histopathology, clinically meaningful reading of a brain MRI is inherently volumetric and context-dependent, requiring signal integration across three planes with patient age, history, and anatomical priors. Vision-language models (VLMs) are now proposed as general-purpose clinical reasoners \cite{moor2023foundation, achiam2023gpt, team2023gemini}, with medical visual question answering (VQA) the dominant evaluation paradigm \cite{lau2018dataset, he2020pathvqa, pal2025rexvqa}. Yet a basic question remains unanswered for 3D neuroimaging. \textit{When a VLM answers a question about a 3D brain MRI scan, is it examining the image, or reciting priors cued by surrounding text}~\cite{asadi2026mirage, agrawal2018don}?

Current medical VQA benchmarks are poorly positioned to separate these two explanations: (1) \textit{Limited dimensionality}; the field is dominated by 2D resources such as VQA-RAD \cite{lau2018dataset}, PathVQA \cite{he2020pathvqa}, and ReXVQA \cite{pal2025rexvqa}, while the few 3D brain MRI efforts are restricted to a single clinical domain \cite{vepa2025multimodal} or are obtained by flattening volumes into 2D slices \cite{wang2024enhancing, peng2025omnibrainbench}. No existing resource exercises volumetric reasoning across diverse neurological conditions. (2) \textit{Unaudited shortcut exploitation}; most benchmarks report headline accuracy without verifying that the image is needed to answer. Asadi et al.\ \cite{asadi2026mirage} show that frontier VLMs retain 70 to 99 percent of their medical VQA accuracy when the image is removed, a result consistent with keyword, cohort-name, and answer-position priors that leak through the text channel \cite{agrawal2018don, goyal2017making}. Accordingly, a leaderboard number overstates image understanding by an unknown margin. (3) \textit{No human visual baseline}; even when shortcut risk is controlled, a low model score is uninterpretable without knowing what a clinician can recover from the same view. Without such a reference, benchmarks conflate a perceptual ceiling with a modeling failure.

\begin{figure}[t]
    \centering
    \includegraphics[width=\textwidth]{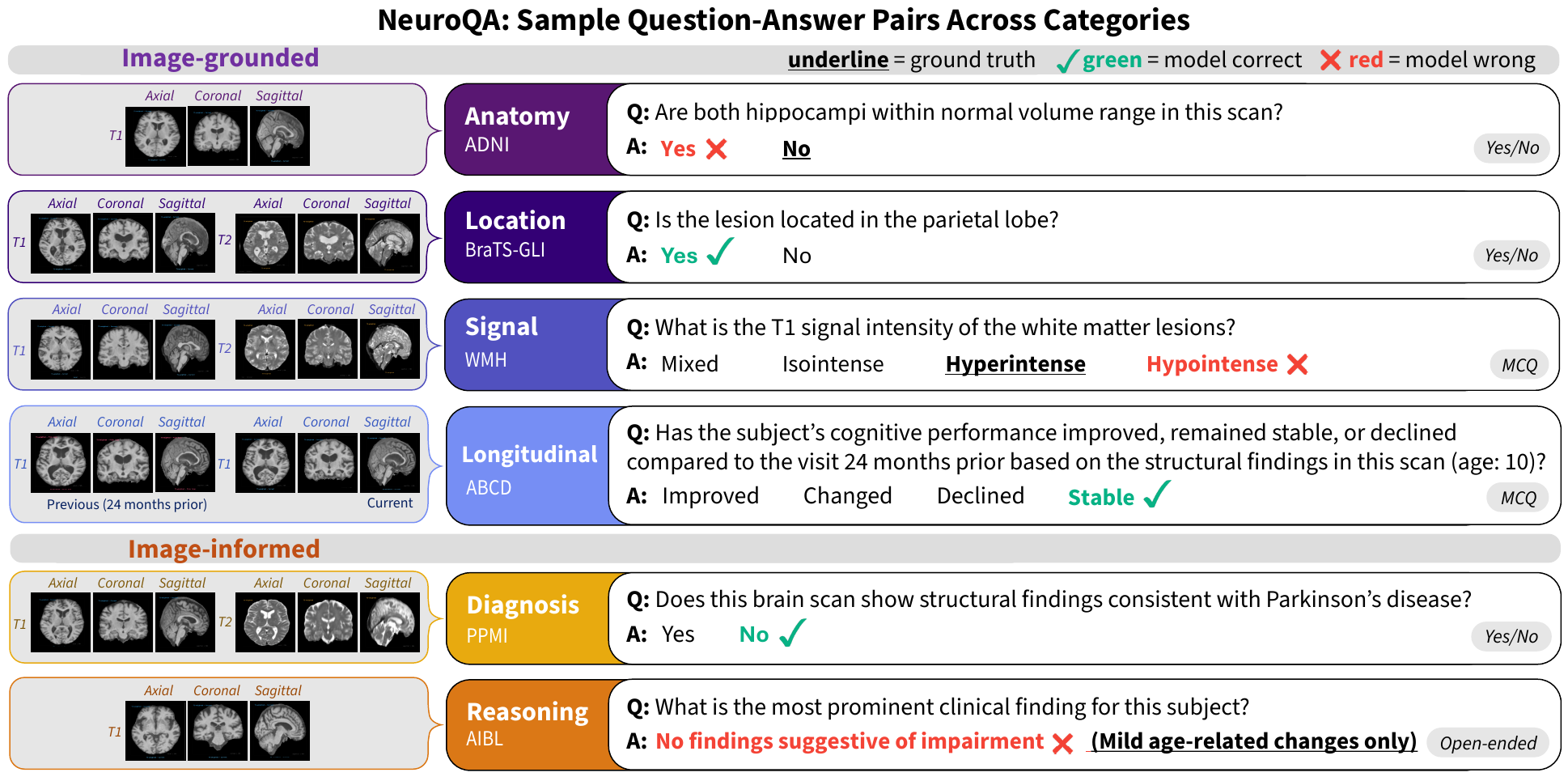}
    \caption{Sample \projectname{} items across six categories, each paired with axial, sagittal, and coronal slices of the 3D volume; T2 is shown alongside T1 where available, and longitudinal items show both prior and current scans. Questions span structural assessment (Anatomy), hemispheric comparison (Location), signal characterization (Signal), temporal change detection (Longitudinal), clinical classification (Diagnosis), and clinical reasoning (Reasoning).}
    \label{fig:samples}
\end{figure}

We introduce \textbf{\projectname{}}, a benchmark for image-conditioned question answering on 3D brain MRI designed from the ground up to address these three gaps (Figure~\ref{fig:samples}). \projectname{} contains \textbf{56{,}953} QA pairs over \textbf{12{,}977} subjects drawn from \textbf{12} neuroimaging datasets, spanning ages 5 to 104 and five clinical domains: Alzheimer's disease, Parkinson's disease, brain tumors, white matter disease, and healthy aging and neurodevelopment. It evaluates 11 clinically grounded reasoning skills across Yes/No, multiple-choice, and open-ended formats, with ground-truth answers derived from FreeSurfer volumetry \cite{desikan2006automated} and structured metadata (9 datasets) or from structured radiology report fields via a negation-aware parser (3 datasets). Importantly, \projectname{} is scoped to a single evaluative claim, namely whether a VLM extracts clinically relevant quantitative information from 3D brain volumes \textit{beyond what text priors or human visual inspection alone recover}. Three assumptions make this claim testable. \textit{(i)} Metadata-derived ground truth is more reproducible than report-derived labels. \textit{(ii)} A text-only baseline near random chance certifies that residual accuracy is image-attributable. \textit{(iii)} A clinician baseline on a 3-plane Neuroimaging Informatics Technology Initiative (NIfTI) viewer bounds what visual inspection alone can recover. The benchmark is not intended to support claims about clinical deployment, prognosis, treatment planning, or report generation.

Our main contributions are summarized as follows:
\begin{itemize}[leftmargin=*, nosep]
    \item We release the largest publicly documented benchmark for 3D brain MRI question answering, with \textbf{56{,}953} QA pairs across 12 datasets, 12{,}977 subjects, 203 templates, 11 categories, and three answer formats. Unlike prior 3D resources, \projectname{} spans multiple clinical domains and preserves volumetric structure end-to-end.
    \item We apply template-level auditing for shortcut-refinement and observe a decrease in text-only accuracy from above 80\% on the raw candidate pool to \textbf{44.6\%}, within 5.1 points of the 39.5\% random-chance floor. Per-category text-only accuracy falls \emph{below} random for Reasoning (33.5\%), Demographic (37.8\%), and Measurement (38.7\%), so residual accuracy on these categories is attributable to image content, not template priors.
    \item We provide clinician visual baselines, with two independent clinicians reading a 3-plane NIfTI viewer reaching closed-format accuracies of 46.7\% and 51.1\% (mean \textbf{48.9\%}) on a 100-item frozen subset, 0.5 percentage points (pp) below the text-only closed floor of 49.4\%. Inter-rater agreement is at the fair-to-moderate boundary (Cohen's $\kappa$~=~0.40), and 23 of 44 disagreements fall in categories where ground truth derives from quantitative metadata (Diagnosis, Severity, Reasoning) or requires precise spatial reasoning across all three planes (Location), consistent with a perceptual ceiling on quantitative-metadata items.
    \item We adopt a two-tier reproducible release where QA pairs are published directly for six public datasets (19{,}262 items), and a deterministic generation pipeline reproduces QA pairs for six DUA-restricted datasets (37{,}691 items). A held-out private test set supports leaderboard evaluation.
\end{itemize}

\section{Related Work}

\subsection{Medical VQA Benchmarks}

Medical VQA benchmarks have been developed for radiology \cite{lau2018dataset, liu2021slake, bae2024mimic, ben2019vqa, ben2021overview}, pathology \cite{he2020pathvqa}, microscopy \cite{li2025microvqa++}, mixed modalities \cite{zhang2023pmc, liu2025gemex, zuo2025medxpertqa, soni2022radqa}, and temporal reasoning \cite{hu2023medicaldiff}. LUMEN \cite{jiang2026lumen} introduced longitudinal VQA for chest X-ray (CXR) prognosis but targets 2D CXR rather than 3D brain MRI. ReXVQA \cite{pal2025rexvqa} is the largest single-modality resource, with 696K multiple-choice items for chest X-ray. These benchmarks operate exclusively on 2D images. None target 3D brain MRI, and none report a text-only baseline alongside overall accuracy to quantify how much performance depends on the image. Table~\ref{tab:comparison} compares \projectname{} with prior benchmarks on six evaluation axes.

\subsection{Shortcut Exploitation in Medical VQA}

VQA models exploit language priors \cite{agrawal2018don, geirhos2020shortcut} and answer-position bias \cite{goyal2017making}. In medical settings, metadata such as age and diagnosis amplifies the effect. Asadi et al. \cite{asadi2026mirage} show that frontier VLMs retain 70 to 99\% of their medical VQA accuracy when images are removed, making medical benchmarks the most affected category. Huang et al. \cite{huang2024visual} find that multimodal systems fabricate visual content when no image is provided, and similar fragility appears across model families \cite{alayrac2022flamingo, radford2021learning, singhal2023large}. These results call for evaluation protocols that separate text-based from image-based performance. \projectname{} enforces this separation through per-template answer balancing, removal of cohort identifiers from question text, and a text-only baseline reported for every category.

\subsection{3D Brain MRI Understanding}

OmniBrainBench \cite{peng2025omnibrainbench} spans 15 brain imaging modalities but projects volumes to 2D slices and does not audit for shortcuts. BrainMD \cite{wang2024enhancing} pairs 2,453 3D scans with reports yet evaluates on representative 2D slices only. mpLLM \cite{vepa2025multimodal} retains full 3D volumes but covers only BraTS tumor data, with roughly 3,000 perturbations of 15 templates, and is released as an evaluation set rather than a public benchmark resource. MM-NeuroOnco \cite{guo2026mm} targets tumor diagnosis alone; 3D-RAD \cite{gai20253d} targets computed tomography (CT) rather than brain MRI. General VLMs \cite{achiam2023gpt, team2023gemini, claude3, liu2023visual, li2023llava, sellergren2025medgemma, tu2024towards, wang2024qwen2, zhang2023biomedclip, nori2023can, dong2024brain, liu2023brainclip,nerrise2026geosae} have been tested on small neuroimaging sets without shortcut analysis or clinician baselines. No existing brain MRI benchmark combines multi-domain coverage, shortcut analysis, and a clinician reference.

\begin{table}[t]
\caption{Comparison with prior medical VQA benchmarks. Formats: answer types (1 = Open, 2 = YN + Open, 3 = YN + MCQ + Open). Audit: dataset-level validation on frozen data (Clinical = clinician rater study, Shortcut = template-level shortcut audit). $^*$3D scans evaluated on representative 2D slices. $^\dagger$Filtered from 2.7M candidates. $^\diamond$Released as an evaluation set, not as a public benchmark resource.}
\label{tab:comparison}
\centering
\setlength{\tabcolsep}{4pt}
\begin{tabular}{lrcccc}
\toprule
Benchmark & QA Pairs & Dim. & Formats & Temporal & Audit \\
\midrule
\multicolumn{6}{l}{\textit{General medical VQA}} \\
VQA-RAD \cite{lau2018dataset} & 3.5K & 2D & 2 & \ding{55} & -- \\
SLAKE \cite{liu2021slake} & 14K & 2D & 2 & \ding{55} & -- \\
PathVQA \cite{he2020pathvqa} & 33K & 2D & 2 & \ding{55} & -- \\
PMC-VQA \cite{zhang2023pmc} & 227K & 2D & 1 & \ding{55} & -- \\
MIMIC-CXR-VQA \cite{bae2024mimic} & 377K & 2D & 2 & \ding{55} & -- \\
ReXVQA \cite{pal2025rexvqa} & 696K & 2D & 1 & \ding{55} & Clinical \\
3D-RAD \cite{gai20253d} & 170K & 3D & 2 & \ding{51} & -- \\
\midrule
\multicolumn{6}{l}{\textit{Brain MRI VQA}} \\
OmniBrain \cite{peng2025omnibrainbench} & 9.5K & 2D & 2 & \ding{55} & -- \\
BrainMD \cite{wang2024enhancing} & 2.5K & 2D$^*$ & 1 & \ding{55} & -- \\
mpLLM \cite{vepa2025multimodal}$^\diamond$ & ${\sim}$3K & 3D & 1 & \ding{55} & -- \\
\midrule
\textbf{\projectname{}} & \textbf{57K}$^\dagger$ & \textbf{3D} & \textbf{3} & \ding{51} & \textbf{Shortcut+Clinical} \\
\bottomrule
\end{tabular}
\end{table}

\section{The \projectname{} Benchmark}
\label{sec:benchmark}

\projectname{} rests on three design commitments. First, every item is paired with a full volumetric scan in Montreal Neurological Institute (MNI-152) space~\cite{fonov2009unbiased} rather than a 2D slice, preserving volumetric structure end-to-end. Second, every ground-truth answer derives from FreeSurfer volumetry~\cite{desikan2006automated, fischl2012freesurfer}, structured metadata, Digital Imaging and Communications in Medicine (DICOM) headers, or a negation-aware parser over RadGenome-Brain MRI reports~\cite{lei2024autorg} for tumor and white-matter datasets, so labels are reproducible and free of reader-level phrasing drift. Third, templates and answer distributions are audited until text-only accuracy falls within 5.1 points of random on the full pool (44.6\% vs.\ 39.5\%), which we use as the design target for shortcut refinement; the corresponding text-only floor on the test-public closed split is 49.4\% against a 39.5\% random chance, which serves as the per-model baseline used in the experiments (\S\ref{sec:experiments}). We detail reasoning categories (\S\ref{sec:categories}), data sources (\S\ref{sec:composition}), the construction pipeline (\S\ref{sec:pipeline}), and summary statistics (\S\ref{sec:stats}).

\subsection{Reasoning Categories}
\label{sec:categories}

\projectname{} organizes its 203 question templates into 11 reasoning categories grouped by the type of evidence required to recover the ground-truth answer. \emph{\ig[Image-grounded]} categories (131 templates) are answerable from direct visual inspection of the 3D volume on a standard 3-plane viewer. \emph{\ii[Image-informed]} categories (72 templates) are answerable only in approximation from the image, and exact labels come from quantitative volumetry or clinical instruments whose output correlates with brain structure but cannot be recovered from visual inspection alone. This split is not cosmetic. It fixes what a clinician-level visual read can recover and therefore the margin a competitive VLM must exceed to demonstrate capability beyond human perception (\S\ref{sec:experiments}). A full template-level breakdown appears in Appendix Table~\ref{tab:grounded}.

\myparagraph{\ig[Image-grounded] categories.}
\textbf{Modality} asks which MRI sequence a scan corresponds to (e.g., T1 vs.\ T2). \textbf{Anatomy} asks about gross morphology, including ventricular size, hippocampal appearance, and cortical thickness relative to age norms. \textbf{Location} asks where a finding lies (lobe, hemisphere, region), requiring integration across axial, sagittal, and coronal planes. \textbf{Measurement} asks for quantitative estimates approximable from orthogonal slices, such as ventricle-to-brain ratio or lesion diameter. \textbf{Signal} asks about lesion intensity on T1 or T2 imaging. \textbf{Longitudinal} asks about structural change between current and prior scans, with both volumes provided. \textbf{Presence} asks whether a specific finding (hippocampal atrophy, ventricular enlargement, white matter hyperintensity) is visible.

\myparagraph{\ii[Image-informed] categories.}
\textbf{Diagnosis} asks which clinical condition is most consistent with the scan, with ground truth taken from dataset-assigned diagnostic labels that reflect multi-modal workup rather than imaging alone. \textbf{Severity} asks for a graded clinical assessment such as Clinical Dementia Rating (CDR) stage~\cite{morris1993clinical} or Unified Parkinson's Disease Rating Scale (UPDRS) motor score~\cite{goetz2008movement}, whose label is anchored to an instrument rather than to voxel statistics. Data-driven analyses nonetheless show that motor-symptom heterogeneity in neurodegenerative cohorts has structure that imaging and movement features can recover~\cite{endo2024data}, which motivates Severity items in addition to Diagnosis. \textbf{Reasoning} asks multi-step questions that combine structural evidence with clinical context, such as which diagnostic category is most consistent with both the degree of hippocampal atrophy and the subject's age. \textbf{Demographic} asks about age or sex, with ground truth taken from the enrollment record, which correlates with but is not determined by brain morphology.

\myparagraph{Answer formats.} Every category is expressible in three formats. Yes/No (YN) items (26{,}378; 46.3\%) hold exact 50/50 answer balance per dataset. Four-option multiple-choice (MCQ) items (19{,}167; 33.7\%) hold uniform 25\% correct-answer positions per dataset. Open-ended free-text items (11{,}408; 20.0\%) admit short phrases and are evaluated under exact match (EM) and token-level F1 (\S\ref{sec:experiments}).

\subsection{Data Sources and Splits}
\label{sec:composition}

\projectname{} assembles 56{,}953 QA pairs over 12{,}977 subjects drawn from 12 neuroimaging datasets spanning ages 5 to 104. The source datasets cover five clinical domains. Alzheimer's disease and mild cognitive impairment are represented by ADNI~\cite{petersen2010alzheimer} and AIBL~\cite{ellis2009australian}. Parkinson's disease is represented by PPMI~\cite{marek2011parkinson}. Primary brain tumors are represented by BraTS-GLI~\cite{baid2021rsna} and BraTS-MEN~\cite{labella20262024}. Small-vessel white matter disease is represented by the White Matter Hyperintensity (WMH) Segmentation Challenge~\cite{kuijf2019standardized} and CC359~\cite{souza2018open}. Healthy aging and neurodevelopment are represented by HCP-YA, HCP-Aging, HCP-Development~\cite{van2013wu}, ABCD~\cite{casey2018adolescent}, and IXI~\cite{hillixi}. Every volume is preprocessed via a standardized sMRI pipeline~\cite{abbasi2025smri} and resampled to MNI-152 space~\cite{fonov2009unbiased} before feature extraction, and clinical metadata is sourced from standardized instruments including CDR~\cite{morris1993clinical}, Montreal Cognitive Assessment (MoCA)~\cite{nasreddine2005montreal}, Mini-Mental State Examination (MMSE)~\cite{folstein1975mini}, NIH Toolbox~\cite{carlozzi2013vi}, Child Behavior Checklist (CBCL)~\cite{achenbach2001manual}, UPDRS~\cite{goetz2008movement}, and Hoehn-Yahr~\cite{hoehn1967parkinsonism}.

\myparagraph{Two-tier release.} Data governance is uneven across these 12 sources, so \projectname{} releases in two tiers. \textbf{Tier~1 (Public)} publishes 19{,}262 QA pairs directly over 2{,}467 subjects from six permissively licensed datasets (BraTS-GLI, BraTS-MEN, WMH, CC359, IXI, HCP-YA). \textbf{Tier~2 (data use agreement, DUA)} publishes a deterministic generation pipeline that reproduces 37{,}691 QA pairs over 10{,}510 subjects bit-for-bit from ADNI, PPMI, ABCD, AIBL, HCP-Aging, and HCP-Development for researchers who obtain their own data access. Both tiers share identical templates, identical validation rules, and identical evaluation code, so any claim established on Tier~1 transfers directly to Tier~2.

\myparagraph{Subject-level partitions.} We partition subjects, not items, into Train (40{,}355 items), Validation (5{,}493), Test-Public (5{,}524), and Test-Private (5{,}581). No subject appears in more than one split, which rules out same-subject leakage by construction. Test-Private is held back for leaderboard submissions with hidden ground truth. The 100-item clinician study subset (\S\ref{sec:experiments}) is drawn exclusively from Test-Public and was frozen prior to rater evaluation. All data were collected under existing data use agreements, and this study was determined not human subjects research by the Institutional Review Board (Protocol \#85709).

\subsection{Construction Pipeline}
\label{sec:pipeline}

\begin{figure}[t]
    \centering
    \includegraphics[width=\textwidth]{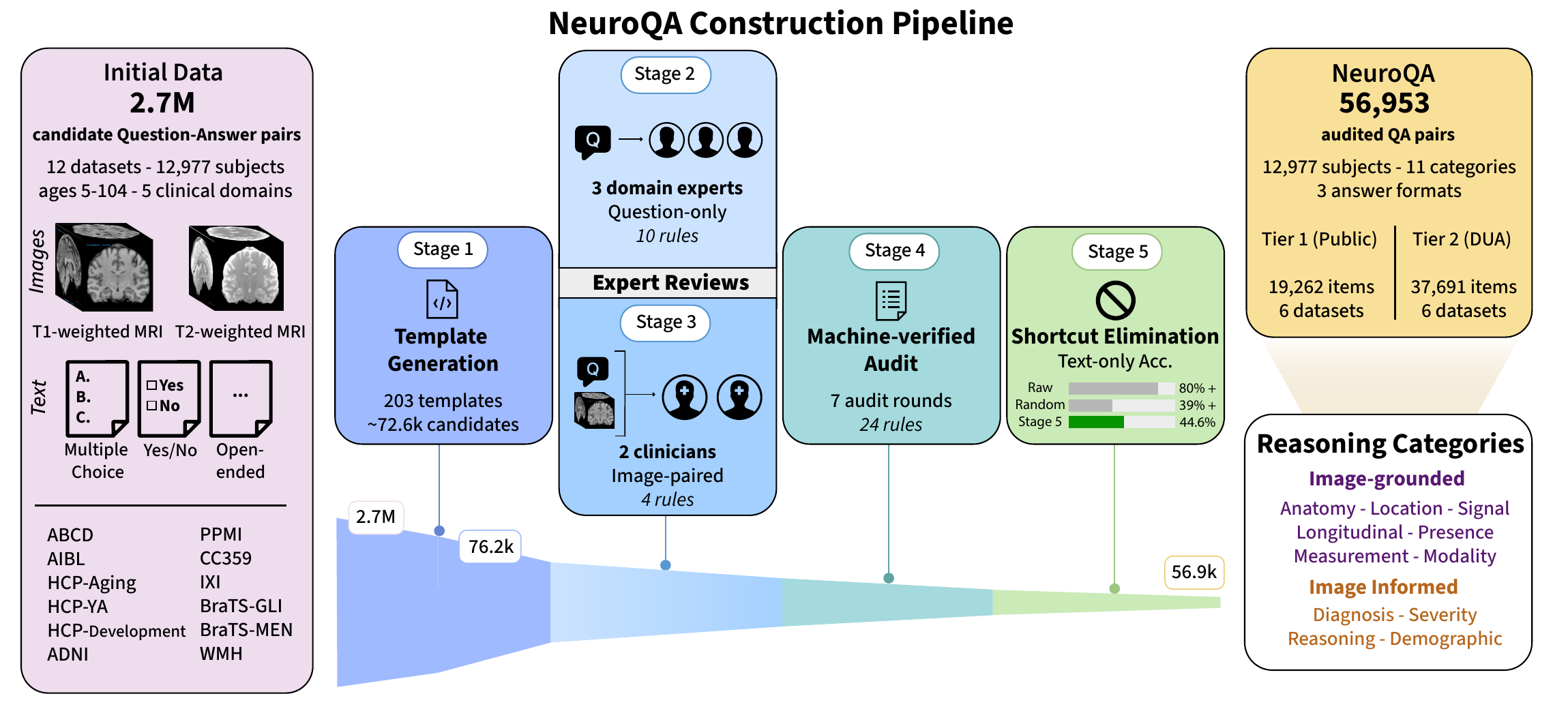}
    \caption{\projectname{} construction pipeline. Five deterministic stages transform raw neuroimaging data from 12 datasets into 56{,}953 validated QA pairs through template-based generation, expert review, machine-verified audits, and shortcut elimination. No LLM is used at any stage, and the full pipeline is reproducible under \texttt{seed=42}.}
    \vspace{-10pt}
    \label{fig:pipeline}
\end{figure}

Figure~\ref{fig:pipeline} shows a five-stage deterministic pipeline (\texttt{seed=42}) reducing 2.7M candidates to 56{,}953 items without LLMs, so Tier~2 reproduction regenerates identical QA files given the same FreeSurfer release. Shortcut elimination is deferred to the final stage to preserve the enforced answer balance.

\myparagraph{Stage 1. Template-based generation.} Twelve dataset-specific scripts emit 72{,}603 pairs from 203 templates. Thresholds for quantitative questions such as ``above-median hippocampal volume'' or ``enlarged ventricles for age'' use cohort-specific medians rather than fixed clinical cutoffs, which keeps the Yes/No prior near 0.5 within each cohort and prevents the template itself from encoding population-level disease priors. Where both a report-derived and a metadata-derived label exist for the same finding, we prefer the metadata-derived one to avoid inheriting radiologist phrasing.

\myparagraph{Stages 2--3. Expert review.} Two rounds of review produce rules R1 through R14. The first round engaged three domain experts on 24 question-only items and surfaced issues of laterality convention, asymmetry thresholds, signal wording, age context, compound MCQ options, and diagnostic-framework ambiguity. The second round engaged two clinicians on 57 image-paired items via the same 3-plane NIfTI viewer later used for the clinician evaluation (\S\ref{sec:experiments}), which surfaced two additional parser corrections covering report-negation handling and full-text signal extraction. All 14 expert rules were applied before any shortcut refinement. Per-item findings appear in Appendix~\ref{app:expert}.

\myparagraph{Stage 4. Machine-verified audits.} Seven audit rounds (R16--R38) verify session alignment, age-bin consistency, BraTS signal cross-validation against parsed reports, removal of dataset identifiers from question text, and image-path integrity. Every rule is a deterministic check with an explicit failure count, so audit output is diffable across runs.

\myparagraph{Stage 5. Shortcut elimination.} The final stage removes templates with $>$95\% answer dominance, downsamples 70--95\% dominance templates (R24), enforces 50/50 YN balance per template per dataset (R23), and redistributes MCQ positions to uniform 25\% (R25). Text-only accuracy on the resulting pool is \textbf{44.6\%}, 5.1 points above the 39.5\% random floor, and falls \emph{below} random on Reasoning (33.5\%), Demographic (37.8\%), and Measurement (38.7\%). The benchmark was frozen after Stage~5; no item was modified after the clinician evaluation (\S\ref{sec:experiments}). Details appear in Appendix~\ref{app:rules}.

\subsection{Dataset Statistics}
\label{sec:stats}

\projectname{} releases 56{,}953 QA pairs across 12{,}977 subjects, 12 neuroimaging datasets, 203 question templates, and 11 reasoning categories, which to our knowledge is the largest publicly documented benchmark for 3D brain MRI question answering to date. Items split into 26{,}378 Yes/No (46.3\%), 19{,}167 four-option MCQ (33.7\%), and 11{,}408 open-ended free-text (20.0\%) pairs. Train, Validation, Test-Public, and Test-Private partitions hold 40{,}355 / 5{,}493 / 5{,}524 / 5{,}581 items respectively, with no cross-split subject leakage. Per-dataset and per-category counts appear in Appendix~\ref{app:dataset}; the full 38-rule audit log appears in Appendix~\ref{app:rules}.

\section{Experiments}
\label{sec:experiments}

\subsection{Evaluation Protocol}

All models are evaluated on test-public (5{,}524 items over 12 datasets and 11 categories), and test-private (5{,}581 items) is reserved for leaderboard submissions with hidden ground truth. We report accuracy overall and per dataset, category, and format, so aggregates cannot mask cohort- or skill-specific failures. Each evaluation item provides the full 3D volume in MNI-152 space (T1-weighted, with T2-weighted included where the \texttt{answerable\_modalities} field permits), the question text, and the four options for MCQ items. Longitudinal items additionally provide the prior scan with its acquisition date. Models may consume the full volume or sample orthogonal 2D slices.

\myparagraph{Answer-format scoring.} YN items use binary accuracy. MCQ items use accuracy after rule-based option extraction on letters A-D. Open-ended items report exact match after normalization and token-level F1. BERTScore~\cite{zhang2019bertscore} is available in the evaluation toolkit but not reported here.

\myparagraph{Shortcut Score.} To separate image-driven from text-driven accuracy, we report a Shortcut Score
\begin{equation}
\text{SS} = 1 - \frac{\text{Acc}_{\text{model}} - \text{Acc}_{\text{text-only}}}{\text{Acc}_{\text{oracle}} - \text{Acc}_{\text{text-only}}}, \qquad \text{Acc}_{\text{oracle}} = 100\%.
\end{equation}
$\text{SS} = 0$ means model accuracy equals oracle, so the model recovers all available signal. $\text{SS} = 1$ means accuracy equals the text-only baseline, indicating reliance on text priors. Intermediate values quantify partial reliance; per-category SS surfaces which reasoning skills gain most from image access. As a worked example, the supervised 3D CNN reaches 43.7\% closed against a 49.4\% text-only floor on test-public, so $\Delta = -5.7$ and $\text{SS} = 1 - (-5.7)/(100 - 49.4) = 1.11$, with $\text{SS} > 1$ indicating below-floor performance. SS is a normalized gap-to-oracle on $[0,1]$ under the text-only floor; values above 1 flag below-floor accuracy but do not by themselves identify shortcut exploitation, since a model that anti-correlates with text priors can also fall below the floor, so SS is reported alongside, not in place of, the image-grounding protocol (\S\ref{sec:grounding}).

\subsection{Baselines}
Table~\ref{tab:baselines} reports baselines on test-public. \textbf{Random} reflects closed-format chance (39.5\%, weighted across YN at 50\% and MCQ at 25\%). \textbf{Text-only} predicts the majority answer per template without image access, yielding a closed floor of 49.4\% on test-public. We report closed-format accuracy (YN\,+\,MCQ) as the primary metric because open-ended scoring (exact match) is not comparable across model architectures; open-ended EM is reported separately in Table~\ref{tab:baselines} and Appendix Table~\ref{tab:open_metrics}. \textbf{Clinician raters} evaluate a 100-item frozen subset on a three-plane NIfTI viewer and achieve a two-rater closed mean of 48.9\%, 0.5\,pp below the text-only floor (\S\ref{sec:clinician_study}). We also include a \textbf{supervised 3D baseline}, a 1.02M-parameter multi-task 3D CNN trained on the train split (40{,}355 items; Appendix~\ref{app:cnn}). It reaches 43.7\% closed ($\Delta = -5.7$, SS\,=\,1.11), below both the text-only and clinician floors, indicating the task is not trivially solvable even with supervised volumetric training. \textbf{Vision-language models} are evaluated zero-shot under the stack condition. Table~\ref{tab:baselines} reports the best-performing model per family from seven families. Gemini-3.1-Pro leads at 47.5\% closed (SS\,=\,1.04), still 1.9\,pp below the text-only floor; GPT-5.2 follows at 46.3\% (SS\,=\,1.06). No model in Table~\ref{tab:baselines} exceeds the text-only or clinician floor, and MedGemma-4B in the broader evaluation (Appendix~\ref{app:vlm}) falls below the 39.5\% random chance, indicating that image input actively degrades performance on some architectures. Per-condition breakdowns, per-category accuracy, and open-ended metrics appear in Appendix Tables~\ref{tab:per_qtype}--\ref{tab:open_metrics}. Results including the private test set are available on the public leaderboard at \url{https://neuroqa.stanford.edu}.

\begin{table}[t]
\caption{Baselines on \projectname{} test-public ($N{=}5{,}524$; $N_{\text{closed}}{=}4{,}454$). Only the best model per VLM family is shown; full per-model results in Fig.~\ref{fig:results} and Tables~\ref{tab:per_qtype}--\ref{tab:open_metrics}. Clinician evaluation is on a 100-item subset (Cohen's $\kappa = 0.40$). \textbf{Random}: analytical chance floor. \textbf{Text-only}: majority answer per template (digits stripped), no image; closed accuracy defines the shortcut floor (49.4\%). $\Delta$: closed accuracy minus text-only floor. SS (Shortcut Score): $1 - \frac{\text{Closed} - 49.4}{50.6}$; SS\,=\,1.00 at text-only, 0.00 at oracle, $>$1.00 = sub-floor. VLMs: stack condition, zero-shot. Open EM is 0.0\% by construction for the Random and Text-only rows because neither strategy reliably exact-matches an open-ended free-text gold without image access.}
\label{tab:baselines}
\centering
\small
\setlength{\tabcolsep}{4pt}
\begin{tabular}{@{}lcccccc@{}}
\toprule
Method & YN & MCQ & Closed & $\Delta$ vs.\ TO & SS & Open EM \\
\midrule
\multicolumn{7}{l}{\textit{Reference floors}} \\
Random (chance)                    & 50.0\% & 25.0\% & 39.5\% & \textcolor{red!70!black}{$-9.9$}  & 1.20 &  0.0\% \\
Text-only (majority per template)  & 57.4\% & 38.1\% & \textbf{49.4\%} & $\pm 0.0$ & 1.00 &  0.0\% \\
\midrule
\multicolumn{7}{l}{\textit{Clinician raters} (100-item subset: 73 YN + 17 MCQ + 10 Open)} \\
Rater 1 (radiology resident)       & 49.3\% & 35.3\% & 46.7\% & \textcolor{red!70!black}{$-2.7$}  & 1.05 & 30.0\% \\
Rater 2 (neurosurgery resident)    & 54.8\% & 35.3\% & \textbf{51.1\%} & \textcolor{green!40!black}{$+1.7$} & \textbf{0.97} & 20.0\% \\
Two-rater mean                     & 52.1\% & 35.3\% & 48.9\% & \textcolor{red!70!black}{$-0.5$}  & 1.01 & 25.0\% \\
\midrule
\multicolumn{7}{l}{\textit{Supervised baseline (trained on \projectname{})}} \\
3D CNN (multi-task)$^\ddagger$     & 56.8\% & 25.5\% & 43.7\% & \textcolor{red!70!black}{$-5.7$}  & 1.11 & 59.6\% \\
\midrule
\multicolumn{7}{l}{\textit{Best vision-language model per family} (zero-shot)} \\
Gemini-3.1-Pro                     & \textbf{58.1\%} & 33.2\% & \textbf{47.5\%} & \textcolor{red!70!black}{$-1.9$} & \textbf{1.04} & 22.9\% \\
GPT-5.2                            & 55.0\% & \textbf{34.6\%} & 46.3\% & \textcolor{red!70!black}{$-3.1$} & 1.06 & 14.2\% \\
GPT-4.1                            & 54.7\% & 33.6\% & 45.7\% & \textcolor{red!70!black}{$-3.7$} & 1.07 & 17.6\% \\
LLaMA-4-Scout                      & 53.1\% & 31.0\% & 43.6\% & \textcolor{red!70!black}{$-5.8$} & 1.11 & \textbf{24.7\%} \\
Claude Sonnet 4.6                  & 51.5\% & 29.3\% & 42.0\% & \textcolor{red!70!black}{$-7.4$} & 1.15 & 21.4\% \\
MedGemma-27B                       & 48.7\% & 26.4\% & 39.2\% & \textcolor{red!70!black}{$-10.2$} & 1.20 & 24.6\% \\
\bottomrule
\multicolumn{7}{l}{\footnotesize $^\ddagger$1.02M-param 3D CNN; Open EM uses closed vocabulary from training answers (Appendix~\ref{app:cnn}).}
\end{tabular}
\end{table}

\begin{figure}[t]
  \centering
  \includegraphics[width=\linewidth]{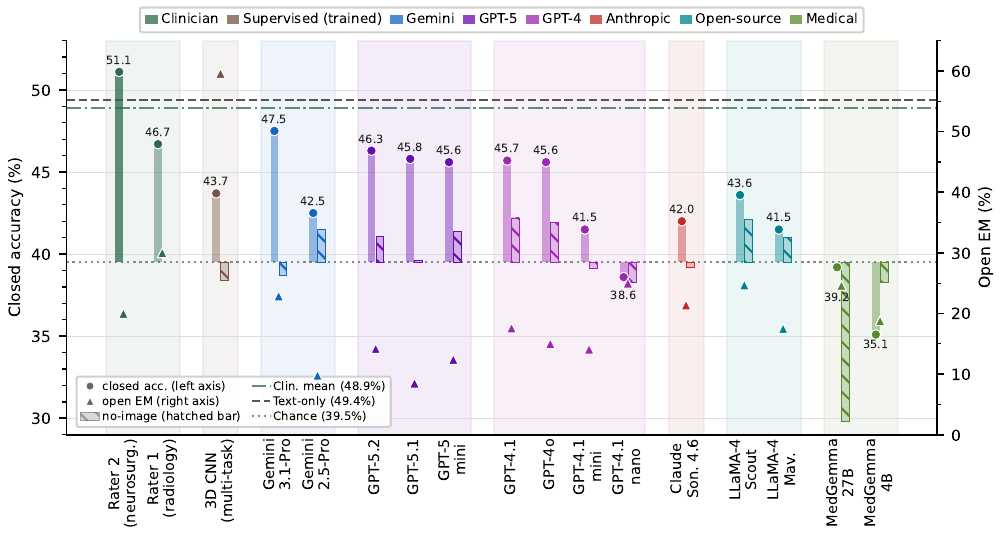}
  \vspace{-10pt}
  \caption{Closed-ended VLM accuracy on test-public ($N{=}5{,}524$, stack condition) versus clinician raters on a frozen 100-item subset. Dotted, dashed, and dash-dot lines mark chance (39.5\%), text-only floor (49.4\%), and clinician mean (48.9\%).}
  \label{fig:results}
\end{figure}

\subsection{Clinician Evaluation}
\label{sec:clinician_study}

Two clinicians (a radiology resident and a neurosurgery resident) evaluated 100 test-public items spanning 100 templates in a web-based 3-plane NIfTI viewer. The viewer displayed T1 and T2 volumes where available; longitudinal items included the prior scan and acquisition date. Raters had no ground truth or feedback, and the benchmark was frozen before evaluation. Rater~1 scored \textbf{45.0\%} and Rater~2 scored \textbf{48.0\%} overall; on closed-format items their accuracies are 46.7\% and 51.1\% (mean \textbf{48.9\%}), 0.5\,pp below the text-only closed floor of 49.4\%. Cohen's $\kappa$~\cite{cohen1960coefficient} is 0.40 (56\% exact agreement).

The category structure supports the \ig/\ii\ split of \S\ref{sec:categories}. Clinicians perform best on directly visible categories (e.g., Modality 100\%/100\%) but plateau on categories whose labels derive from FreeSurfer volumetry or clinical instruments (Reasoning 28.6\%/28.6\%; Severity 28.6\%/42.9\%). Item-weighted across all formats, they average 50.0\% on 62 \ig\ items and 42.1\% on 38 \ii\ items, explaining why the aggregate closed mean sits below the text-only floor. On the 56 agreed items, only 27 are correct against ground truth, so the remaining shared errors indicate a systematic perceptual ceiling on \ii\ items rather than rater noise. A VLM must therefore exceed the text-only floor with gains concentrated on \ii\ categories to show extraction beyond template priors and clinician visual inspection; no current VLM clears this bar. Appendix~\ref{app:clinician_analysis} gives the per-category analysis.

\begin{table}[t]
\vspace{-10pt}
\caption{Per-category clinician accuracy on the 100-item frozen subset. Closed-format means appear in Table~\ref{tab:baselines}; detailed interpretation is in Appendix~\ref{app:clinician_analysis}. Image-grounded subtotal (62 items): Rater 1 48.4\%, Rater 2 51.6\%, mean 50.0\%. Image-informed subtotal (38 items): Rater 1 39.5\%, Rater 2 44.7\%, mean 42.1\%.}
\label{tab:clinician}
\centering
\small
\begin{tabular}{lcccl}
\toprule
Category & Rater 1 & Rater 2 & Agreement & Type \\
\midrule
Modality & 100.0\% (2/2) & 100.0\% (2/2) & 100\% & \ig[Image-grounded] \\
Measurement & 100.0\% (2/2) & 50.0\% (1/2) & 50\% & \ig[Image-grounded] \\
Anatomy & 55.6\% (10/18) & 61.1\% (11/18) & 67\% & \ig[Image-grounded] \\
Longitudinal & 57.1\% (4/7) & 57.1\% (4/7) & 57\% & \ig[Image-grounded] \\
Signal & 57.1\% (4/7) & 57.1\% (4/7) & 57\% & \ig[Image-grounded] \\
Location & 27.3\% (3/11) & 54.5\% (6/11) & 36\% & \ig[Image-grounded] \\
Presence & 33.3\% (5/15) & 26.7\% (4/15) & 73\% & \ig[Image-grounded] \\
\midrule
Demographic & 66.7\% (4/6) & 33.3\% (2/6) & 33\% & \ii[Image-informed] \\
Diagnosis & 45.5\% (5/11) & 72.7\% (8/11) & 18\% & \ii[Image-informed] \\
Severity & 28.6\% (2/7) & 42.9\% (3/7) & 43\% & \ii[Image-informed] \\
Reasoning & 28.6\% (4/14) & 28.6\% (4/14) & 79\% & \ii[Image-informed] \\
\midrule
\textbf{Overall} & \textbf{45.0\%} & \textbf{48.0\%} & \textbf{56\%} ($\kappa = 0.40$) & \textbf{All-format mean: 46.5\%} \\
\bottomrule
\end{tabular}
\end{table}

\subsection{Shortcut Resistance Across Benchmarks}

\citet{asadi2026mirage} report that frontier VLMs retain 70 to 99\% of medical VQA accuracy with images removed, revealing textual shortcuts in benchmark structure. Our Stage-5 refinement removes highly dominant templates, downsamples weaker imbalances, enforces 50/50 Yes/No balance and uniform MCQ positions per dataset, and strips dataset identifiers from question text. Text-only accuracy falls to 44.6\%, only \textbf{+5.1\,pp} above random and below random on Reasoning, Demographic, and Measurement. Table~\ref{tab:shortcut_comparison} compares this margin against public closed-format medical VQA benchmarks using the higher of Gemini-2.5-Pro text-only accuracy and the majority-per-template baseline; \projectname{} has the lowest shortcut margin among the benchmarks tested.

\begin{table}[t]
\caption{Shortcut resistance of public closed-format medical VQA benchmarks. Image-free accuracy is the maximum of Gemini-2.5-Pro text-only accuracy and the majority-per-template baseline; CIs are 95\% Wilson intervals. ReXVQA is gated, PMC-VQA/SLAKE are predominantly open-ended, and BrainMD/mpLLM have n${<}$5K closed items.}
\label{tab:shortcut_comparison}
\centering
\small
\setlength{\tabcolsep}{4pt}
\begin{tabular}{lcrccc}
\toprule
Benchmark & Domain & $n$ & Random & Image-free \scriptsize{[$95\%$ CI]} & Shortcut margin \\
\midrule
PathVQA (YN)~\cite{he2020pathvqa}            & 2D pathology   & 3{,}362   & $50.0\%$ & $63.3$ \scriptsize{[$61.7,\,64.9$]}  & $+13.3$\,pp \\
3D-RAD~\cite{gai20253d}                    & 3D chest CT    & 29{,}218  & $45.1\%$ & $68.6$ \scriptsize{[$68.0,\,69.1$]}  & $+23.5$\,pp \\
MedBookVQA~\cite{yip2025medbookvqa}           & medical textbooks & 5{,}000 & $25.0\%$ & $49.2$ \scriptsize{[$47.8,\,50.6$]}  & $+24.2$\,pp \\
MedXpertQA-MM~\cite{zuo2025medxpertqa}        & expert exam    & 2{,}000   & $20.0\%$ & $46.7$ \scriptsize{[$44.5,\,48.9$]}  & $+26.7$\,pp \\
M3D-VQA~\cite{bai2024m3d}                     & 3D mixed       & 5{,}000   & $25.0\%$ & $52.0$ \scriptsize{[$50.6,\,53.4$]}  & $+27.0$\,pp \\
MMMU H\&M~\cite{yue2024mmmu}                  & exam-style mixed & 1{,}857 & $26.0\%$ & $55.1$ \scriptsize{[$52.8,\,57.3$]}  & $+29.1$\,pp \\
\midrule
\textbf{\projectname{}} & \textbf{3D brain MRI} & \textbf{45{,}545} & $\mathbf{39.5\%}$ & $\mathbf{44.6}$ \scriptsize{$\mathbf{[44.1,\,45.1]}$} & $\mathbf{+5.1}$\,\textbf{pp} \\
\bottomrule
\end{tabular}
\end{table}


\subsection{Image-Grounding Protocol}
\label{sec:grounding}

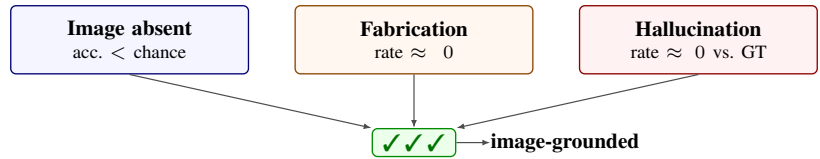
\begin{figure}[H]
\centering
\begin{tikzpicture}[
  font=\footnotesize,
  testbox/.style={
    draw, rounded corners=2pt, line width=0.5pt,
    text width=0.21\linewidth, minimum height=9mm,
    align=center, inner sep=3pt
  },
  blueb/.style={testbox, draw=blue!45!black, fill=blue!4},
  orangeb/.style={testbox, draw=orange!55!black, fill=orange!6},
  redb/.style={testbox, draw=red!55!black, fill=red!5},
  arrow/.style={-{Latex[length=3pt]}, line width=0.45pt, draw=gray!60!black}
]

\node[blueb] (a) {%
  \textbf{Image absent}\\[-1pt]
  \scriptsize acc.\ $<$ chance};

\node[orangeb, right=6mm of a] (b) {%
  \textbf{Fabrication}\\[-1pt]
  \scriptsize rate $\approx 0$};

\node[redb, right=6mm of b] (c) {%
  \textbf{Hallucination}\\[-1pt]
  \scriptsize rate $\approx 0$ vs.\ GT};

\path (a.south) -- (c.south) coordinate[midway] (mid);

\node[draw=green!55!black, fill=green!8, line width=0.5pt,
      rounded corners=2pt, inner sep=2pt, inner xsep=4pt,
      font=\small] (and) at ($(mid)+(0,-9mm)$)
  {\textcolor{green!45!black}{\ding{51}\,\ding{51}\,\ding{51}}};

\node[right=4.5mm of and, font=\footnotesize, inner sep=0pt] (yes)
  {\textbf{image-grounded}};

\draw[arrow] (a.south) -- (and.north west);
\draw[arrow] (b.south) -- (and.north);
\draw[arrow] (c.south) -- (and.north east);
\draw[arrow] (and.east) -- (yes.west);

\end{tikzpicture}
\caption{Image-grounding protocol. A VLM is treated as image-grounded only when all three stress tests pass. \emph{Image absent}: closed accuracy with the image withheld falls toward chance. \emph{Fabrication}: confident visual claims without an image stay near zero. \emph{Hallucination}: visual claims contradicting FreeSurfer ground truth stay low. Failure on any test indicates text-prior reliance rather than 3D image understanding.}
\vspace{-10pt}
\label{fig:grounding}
\end{figure}

Beyond closed accuracy, \projectname{} defines an image-grounding protocol so submissions can certify that gains reflect 3D image understanding rather than text priors. The protocol comprises three stress tests (Figure~\ref{fig:grounding}): \emph{image-absent accuracy} (image withheld), \emph{fabrication rate} (confident visual descriptions produced without an image), and \emph{hallucination rate} (visual claims contradicting the image, verified against FreeSurfer ground truth)~\cite{asadi2026mirage, geirhos2020shortcut, huang2024visual}. A closed accuracy counts as image-grounded only when all three tests pass. Per-model protocol results are tracked on the public leaderboard alongside accuracy.

\section{Conclusion}

We introduced \projectname{}, a shortcut-audited 3D brain MRI benchmark with 56{,}953 QA pairs across 12 datasets, 203 templates, and 11 reasoning categories. Its deterministic pipeline drives text-only accuracy to 44.6\%, while a two-rater clinician study places the human visual closed reference at 48.9\%, 0.5\,pp below the text-only floor. Together, these define the floor a VLM must exceed to demonstrate 3D image understanding beyond template priors and clinician visual inspection. No zero-shot VLM in Table~\ref{tab:baselines} clears this floor, but the gap to the 100\% oracle and the clinician profile (50.0\% on \ig\ items and 42.1\% on \ii\ items where ground truth is not visually recoverable) leave the benchmark solvable in principle by models that extract quantitative information from the 3D volume beyond visual inspection. \textbf{Tier~1 (Public)} QA pairs, \textbf{Tier~2 (DUA-restricted)} generation scripts, hidden test-private evaluation, and the image-grounding protocols make this claim reproducible and testable for groups with standard neuroimaging access. The broader impact is a more transparent evaluation culture for medical VLMs, where claims about neuroimaging capability must survive shortcut, fabrication, and hallucination checks.

\myparagraph{Limitations and future work.} \projectname{} is designed as an evaluation benchmark and not for clinical deployment. Future releases will address white matter hyperintensity T1 items tagged for T2/FLAIR-aware scoring, visual assessment quality control, and additional, higher-level, clinical expertise. 

\begin{ack}
This research was supported in part by the National Institutes of Health grant numbers AG089169, AG084471, AG073107, AG072701, and AG047366. The content is solely the responsibility of the authors and does not necessarily represent the official views of the NIH. This study
was also supported by the Stanford University Institute for Human-Centered AI (HAI) Hoffman-Yee Award.
\end{ack}


{
\small
\bibliographystyle{unsrtnat}
\bibliography{references}

@article{lau2018dataset,
  title={A dataset of clinically generated visual questions and answers about radiology images},
  author={Lau, Jason J and Gayen, Soumya and Ben Abacha, Asma and Demner-Fushman, Dina},
  journal={Scientific data},
  volume={5},
  number={1},
  pages={180251},
  year={2018},
  publisher={Nature Publishing Group}
}

@article{he2020pathvqa,
  title={Pathvqa: 30000+ questions for medical visual question answering},
  author={He, Xuehai and Zhang, Yichen and Mou, Luntian and Xing, Eric and Xie, Pengtao},
  journal={arXiv preprint arXiv:2003.10286},
  year={2020}
}

@inproceedings{liu2021slake,
  title={Slake: A semantically-labeled knowledge-enhanced dataset for medical visual question answering},
  author={Liu, Bo and Zhan, Li-Ming and Xu, Li and Ma, Lin and Yang, Yan and Wu, Xiao-Ming},
  booktitle={2021 IEEE 18th international symposium on biomedical imaging (ISBI)},
  pages={1650--1654},
  year={2021},
  organization={IEEE}
}

@article{zhang2023pmc,
  title={Pmc-vqa: Visual instruction tuning for medical visual question answering},
  author={Zhang, Xiaoman and Wu, Chaoyi and Zhao, Ziheng and Lin, Weixiong and Zhang, Ya and Wang, Yanfeng and Xie, Weidi},
  journal={arXiv preprint arXiv:2305.10415},
  year={2023}
}

@article{bae2024mimic,
  title={MIMIC-Ext-MIMIC-CXR-VQA: a complex, diverse, and large-scale visual question answering dataset for chest X-ray images},
  author={Bae, Seongsu and Kyung, Daeun and Ryu, Jaehee and Cho, Eunbyeol and Lee, Gyubok and Kweon, Sunjun and Oh, Jungwoo and JI, Lei and Chang, Eric and Kim, Tackeun and others},
  journal={PhysioNet},
  year={2024}
}

@inproceedings{pal2025rexvqa,
  title={Rexvqa: A large-scale visual question answering benchmark for generalist chest x-ray understanding},
  author={Pal, Ankit and Lee, Jung-Oh and Zhang, Xiaoman and Sankarasubbu, Malaikannan and Roh, Seunghyeon and Kim, Won Jung and Lee, Meesun and Rajpurkar, Pranav},
  booktitle={Biocomputing 2026: Proceedings of the Pacific Symposium},
  pages={251--264},
  year={2025},
  organization={World Scientific}
}

@article{peng2025omnibrainbench,
  title={OmniBrainBench: A Comprehensive Multimodal Benchmark for Brain Imaging Analysis Across Multi-stage Clinical Tasks},
  author={Peng, Zhihao and Wang, Cheng and Liu, Shengyuan and Liang, Zhiying and Ye, Zanting and Ju, Minjie and Woo, PeterYM and Yuan, Yixuan},
  journal={arXiv preprint arXiv:2511.00846},
  year={2025}
}

@inproceedings{ben2021overview,
  title={Overview of the vqa-med task at imageclef 2021: Visual question answering and generation in the medical domain},
  author={Ben Abacha, Asma and Sarrouti, Mourad and Demner-Fushman, Dina and Hasan, Sadid A and M{\"u}ller, Henning},
  booktitle={Proceedings of the CLEF 2021 Conference and Labs of the Evaluation Forum-working notes},
  year={2021},
  organization={21-24 September 2021}
}

@article{hu2023medicaldiff,
  title={Medicaldiff-vqa: a large-scale medical dataset for difference visual question answering on chest x-ray images},
  author={Hu, Xinyue and Gu, Lin and An, Qiyuan and Zhang, Mengliang and Liu, Liangchen and Kobayashi, Kazuma and Harada, Tatsuya and Summers, R and Zhu, Yingying},
  journal={PhysioNet},
  volume={12},
  pages={13},
  year={2023}
}

@inproceedings{liu2025gemex,
  title={Gemex: A large-scale, groundable, and explainable medical vqa benchmark for chest x-ray diagnosis},
  author={Liu, Bo and Zou, Ke and Zhan, Li-Ming and Lu, Zexin and Dong, Xiaoyu and Chen, Yidi and Xie, Chengqiang and Cao, Jiannong and Wu, Xiao-Ming and Fu, Huazhu},
  booktitle={Proceedings of the IEEE/CVF International Conference on Computer Vision},
  pages={21310--21320},
  year={2025}
}

@inproceedings{agrawal2018don,
  title={Don't just assume; look and answer: Overcoming priors for visual question answering},
  author={Agrawal, Aishwarya and Batra, Dhruv and Parikh, Devi and Kembhavi, Aniruddha},
  booktitle={Proceedings of the IEEE conference on computer vision and pattern recognition},
  pages={4971--4980},
  year={2018}
}

@inproceedings{goyal2017making,
  title={Making the v in vqa matter: Elevating the role of image understanding in visual question answering},
  author={Goyal, Yash and Khot, Tejas and Summers-Stay, Douglas and Batra, Dhruv and Parikh, Devi},
  booktitle={Proceedings of the IEEE conference on computer vision and pattern recognition},
  pages={6904--6913},
  year={2017}
}

@article{geirhos2020shortcut,
  title={Shortcut learning in deep neural networks},
  author={Geirhos, Robert and Jacobsen, J{\"o}rn-Henrik and Michaelis, Claudio and Zemel, Richard and Brendel, Wieland and Bethge, Matthias and Wichmann, Felix A},
  journal={Nature Machine Intelligence},
  volume={2},
  number={11},
  pages={665--673},
  year={2020},
  publisher={Nature Publishing Group UK London}
}

@article{moor2023foundation,
  title={Foundation models for generalist medical artificial intelligence},
  author={Moor, Michael and Banerjee, Oishi and Abad, Zahra Shakeri Hossein and Krumholz, Harlan M and Leskovec, Jure and Topol, Eric J and Rajpurkar, Pranav},
  journal={Nature},
  volume={616},
  number={7956},
  pages={259--265},
  year={2023},
  publisher={Nature Publishing Group UK London}
}

@article{achiam2023gpt,
  title={Gpt-4 technical report},
  author={Achiam, Josh and Adler, Steven and Agarwal, Sandhini and Ahmad, Lama and Akkaya, Ilge and Aleman, Florencia Leoni and Almeida, Diogo and Altenschmidt, Janko and Altman, Sam and Anadkat, Shyamal and others},
  journal={arXiv preprint arXiv:2303.08774},
  year={2023}
}

@article{li2023llava,
  title={Llava-med: Training a large language-and-vision assistant for biomedicine in one day},
  author={Li, Chunyuan and Wong, Cliff and Zhang, Sheng and Usuyama, Naoto and Liu, Haotian and Yang, Jianwei and Naumann, Tristan and Poon, Hoifung and Gao, Jianfeng},
  journal={Advances in Neural Information Processing Systems},
  volume={36},
  pages={28541--28564},
  year={2023}
}

@article{tu2024towards,
  title={Towards generalist biomedical AI},
  author={Tu, Tao and Azizi, Shekoofeh and Driess, Danny and Schaekermann, Mike and Amin, Mohamed and Chang, Pi-Chuan and Carroll, Andrew and Lau, Charles and Tanno, Ryutaro and Ktena, Ira and others},
  journal={Nejm Ai},
  volume={1},
  number={3},
  pages={AIoa2300138},
  year={2024},
  publisher={Massachusetts Medical Society}
}

@article{sellergren2025medgemma,
  title={Medgemma technical report},
  author={Sellergren, Andrew and Kazemzadeh, Sahar and Jaroensri, Tiam and Kiraly, Atilla and Traverse, Madeleine and Kohlberger, Timo and Xu, Shawn and Jamil, Fayaz and Hughes, C{\'\i}an and Lau, Charles and others},
  journal={arXiv preprint arXiv:2507.05201},
  year={2025}
}

@article{wang2024qwen2,
  title={Qwen2-vl: Enhancing vision-language model's perception of the world at any resolution},
  author={Wang, Peng and Bai, Shuai and Tan, Sinan and Wang, Shijie and Fan, Zhihao and Bai, Jinze and Chen, Keqin and Liu, Xuejing and Wang, Jialin and Ge, Wenbin and others},
  journal={arXiv preprint arXiv:2409.12191},
  year={2024}
}

@article{dong2024brain,
  title={Brain-jepa: Brain dynamics foundation model with gradient positioning and spatiotemporal masking},
  author={Dong, Zijian and Li, Ruilin and Wu, Yilei and Nguyen, Thuan T and Chong, Joanna S and Ji, Fang and Tong, Nathanael R and Chen, Christopher L and Zhou, Juan H},
  journal={Advances in Neural Information Processing Systems},
  volume={37},
  pages={86048--86073},
  year={2024}
}

@article{liu2023brainclip,
  title={Brainclip: Bridging brain and visual-linguistic representation via clip for generic natural visual stimulus decoding},
  author={Liu, Yulong and Ma, Yongqiang and Zhou, Wei and Zhu, Guibo and Zheng, Nanning},
  journal={arXiv preprint arXiv:2302.12971},
  year={2023}
}

@article{petersen2010alzheimer,
  title={Alzheimer's disease Neuroimaging Initiative (ADNI) clinical characterization},
  author={Petersen, Ronald Carl and Aisen, Paul S and Beckett, Laurel A and Donohue, Michael C and Gamst, Anthony Collins and Harvey, Danielle J and Jack Jr, Clifford R and Jagust, William J and Shaw, Leslie M and Toga, Arthur W and others},
  journal={Neurology},
  volume={74},
  number={3},
  pages={201--209},
  year={2010},
  publisher={Lippincott Williams \& Wilkins}
}

@article{marek2011parkinson,
  title={The Parkinson progression marker initiative (PPMI)},
  author={Marek, Kenneth and Jennings, Danna and Lasch, Shirley and Siderowf, Andrew and Tanner, Caroline and Simuni, Tanya and Coffey, Chris and Kieburtz, Karl and Flagg, Emily and Chowdhury, Sohini and others},
  journal={Progress in neurobiology},
  volume={95},
  number={4},
  pages={629--635},
  year={2011},
  publisher={Elsevier}
}

@article{casey2018adolescent,
  title={The adolescent brain cognitive development (ABCD) study: imaging acquisition across 21 sites},
  author={Casey, Betty Jo and Cannonier, Tariq and Conley, May I and Cohen, Alexandra O and Barch, Deanna M and Heitzeg, Mary M and Soules, Mary E and Teslovich, Theresa and Dellarco, Danielle V and Garavan, Hugh and others},
  journal={Developmental cognitive neuroscience},
  volume={32},
  pages={43--54},
  year={2018},
  publisher={Elsevier}
}

@article{van2013wu,
  title={The WU-Minn human connectome project: an overview},
  author={Van Essen, David C and Smith, Stephen M and Barch, Deanna M and Behrens, Timothy EJ and Yacoub, Essa and Ugurbil, Kamil and Wu-Minn HCP Consortium and others},
  journal={Neuroimage},
  volume={80},
  pages={62--79},
  year={2013},
  publisher={Elsevier}
}

@article{ellis2009australian,
  title={The Australian Imaging, Biomarkers and Lifestyle (AIBL) study of aging: methodology and baseline characteristics of 1112 individuals recruited for a longitudinal study of Alzheimer's disease},
  author={Ellis, Kathryn A and Bush, Ashley I and Darby, David and De Fazio, Daniela and Foster, Jonathan and Hudson, Peter and Lautenschlager, Nicola T and Lenzo, Nat and Martins, Ralph N and Maruff, Paul and others},
  journal={International psychogeriatrics},
  volume={21},
  number={4},
  pages={672--687},
  year={2009},
  publisher={Cambridge University Press}
}

@article{baid2021rsna,
  title={The rsna-asnr-miccai brats 2021 benchmark on brain tumor segmentation and radiogenomic classification},
  author={Baid, Ujjwal and Ghodasara, Satyam and Mohan, Suyash and Bilello, Michel and Calabrese, Evan and Colak, Errol and Farahani, Keyvan and Kalpathy-Cramer, Jayashree and Kitamura, Felipe C and Pati, Sarthak and others},
  journal={arXiv preprint arXiv:2107.02314},
  year={2021}
}

@article{labella20262024,
  title={The 2024 Brain Tumor Segmentation Challenge Meningioma Radiotherapy (BraTS-MEN-RT) dataset},
  author={LaBella, Dominic and Schumacher, Katherine and Mix, Michael and Leu, Kevin and McBurney-Lin, Shan and Nedelec, Pierre and Villanueva-Meyer, Javier and Raleigh, David R and Shapey, Jonathan and Vercauteren, Tom and others},
  journal={Scientific Data},
  year={2026},
  publisher={Nature Publishing Group UK London}
}

@article{souza2018open,
  title={An open, multi-vendor, multi-field-strength brain MR dataset and analysis of publicly available skull stripping methods agreement},
  author={Souza, Roberto and Lucena, Oeslle and Garrafa, Julia and Gobbi, David and Saluzzi, Marina and Appenzeller, Simone and Rittner, Let{\'\i}cia and Frayne, Richard and Lotufo, Roberto},
  journal={NeuroImage},
  volume={170},
  pages={482--494},
  year={2018},
  publisher={Elsevier}
}

@misc{hillixi,
  title={IXI dataset-Information eXtraction from Images project (EPSRC GR/S21533/02)[Internet]. 2006 [cited 2013 May 7]},
  author={Hill, D and Williams, S and Hawkes, D and Smith, SM},
  year={2006}
}

@article{kuijf2019standardized,
  title={Standardized assessment of automatic segmentation of white matter hyperintensities and results of the WMH segmentation challenge},
  author={Kuijf, Hugo J and Biesbroek, J Matthijs and De Bresser, Jeroen and Heinen, Rutger and Andermatt, Simon and Bento, Mariana and Berseth, Matt and Belyaev, Mikhail and Cardoso, M Jorge and Casamitjana, Adria and others},
  journal={IEEE transactions on medical imaging},
  volume={38},
  number={11},
  pages={2556--2568},
  year={2019},
  publisher={IEEE}
}

@article{desikan2006automated,
  title={An automated labeling system for subdividing the human cerebral cortex on MRI scans into gyral based regions of interest},
  author={Desikan, Rahul S and S{\'e}gonne, Florent and Fischl, Bruce and Quinn, Brian T and Dickerson, Bradford C and Blacker, Deborah and Buckner, Randy L and Dale, Anders M and Maguire, R Paul and Hyman, Bradley T and others},
  journal={Neuroimage},
  volume={31},
  number={3},
  pages={968--980},
  year={2006},
  publisher={Elsevier}
}

@article{fischl2012freesurfer,
  title={FreeSurfer},
  author={Fischl, Bruce},
  journal={Neuroimage},
  volume={62},
  number={2},
  pages={774--781},
  year={2012},
  publisher={Elsevier}
}

@article{fonov2009unbiased,
  title={Unbiased nonlinear average age-appropriate brain templates from birth to adulthood},
  author={Fonov, Vladimir S and Evans, Alan C and McKinstry, Robert C and Almli, C Robert and Collins, DL},
  journal={NeuroImage},
  volume={47},
  pages={S102},
  year={2009},
  publisher={Elsevier}
}

@article{nasreddine2005montreal,
  title={The Montreal Cognitive Assessment, MoCA: a brief screening tool for mild cognitive impairment},
  author={Nasreddine, Ziad S and Phillips, Natalie A and B{\'e}dirian, Val{\'e}rie and Charbonneau, Simon and Whitehead, Victor and Collin, Isabelle and Cummings, Jeffrey L and Chertkow, Howard},
  journal={Journal of the American Geriatrics Society},
  volume={53},
  number={4},
  pages={695--699},
  year={2005},
  publisher={Wiley Online Library}
}

@article{folstein1975mini,
  title={Mini-mental state},
  author={Folstein, Marshal F and Folstein, Susan E and McHugh, Paul R},
  journal={Journal of psychiatric research},
  volume={12},
  number={3},
  pages={189--198},
  year={1975}
}

@article{goetz2008movement,
  title={Movement Disorder Society-sponsored revision of the Unified Parkinson's Disease Rating Scale (MDS-UPDRS): scale presentation and clinimetric testing results},
  author={Goetz, Christopher G and Tilley, Barbara C and Shaftman, Stephanie R and Stebbins, Glenn T and Fahn, Stanley and Martinez-Martin, Pablo and Poewe, Werner and Sampaio, Cristina and Stern, Matthew B and Dodel, Richard and others},
  journal={Movement disorders: official journal of the Movement Disorder Society},
  volume={23},
  number={15},
  pages={2129--2170},
  year={2008},
  publisher={Wiley Online Library}
}

@article{hoehn1967parkinsonism,
  title={Parkinsonism: onset, progression, and mortality},
  author={Hoehn, Margaret M and Yahr, Melvin D},
  journal={Neurology},
  volume={17},
  number={5},
  pages={427--427},
  year={1967},
  publisher={Lippincott Williams \& Wilkins}
}

@article{morris1993clinical,
  title={The Clinical Dementia Rating (CDR) current version and scoring rules},
  author={Morris, John C},
  journal={Neurology},
  volume={43},
  number={11},
  pages={2412--2412},
  year={1993},
  publisher={Lippincott Williams \& Wilkins}
}

@article{carlozzi2013vi,
  title={VI. NIH Toolbox Cognition Battery (CB): measuring processing speed},
  author={Carlozzi, Noelle E and Tulsky, David S and Kail, Robert V and Beaumont, Jennifer L},
  journal={Monographs of the Society for Research in Child Development},
  volume={78},
  number={4},
  pages={88--102},
  year={2013},
  publisher={Oxford University Press}
}

@article{achenbach2001manual,
  title={Manual for the ASEBA school-age forms and profiles},
  author={Achenbach, Thomas M},
  journal={University of Vermont Research Center for Children, Youth, and Families},
  year={2001}
}

@article{zhang2019bertscore,
  title={Bertscore: Evaluating text generation with bert},
  author={Zhang, Tianyi and Kishore, Varsha and Wu, Felix and Weinberger, Kilian Q and Artzi, Yoav},
  journal={arXiv preprint arXiv:1904.09675},
  year={2019}
}

@article{asadi2026mirage,
  title={Mirage the illusion of visual understanding},
  author={Asadi, Mohammad and O'Sullivan, Jack W and Cao, Fang and Nedaee, Tahoura and Fardi, Kamyar and Li, Fei-Fei and Adeli, Ehsan and Ashley, Euan},
  journal={arXiv preprint arXiv:2603.21687},
  year={2026}
}

@article{zuo2025medxpertqa,
  title={Medxpertqa: Benchmarking expert-level medical reasoning and understanding},
  author={Zuo, Yuxin and Qu, Shang and Li, Yifei and Chen, Zhangren and Zhu, Xuekai and Hua, Ermo and Zhang, Kaiyan and Ding, Ning and Zhou, Bowen},
  journal={arXiv preprint arXiv:2501.18362},
  year={2025}
}

@article{li2025microvqa++,
  title={MicroVQA++: High-Quality Microscopy Reasoning Dataset with Weakly Supervised Graphs for Multimodal Large Language Model},
  author={Li, Manyu and He, Ruian and Ma, Chenxi and Tan, Weimin and Yan, Bo},
  journal={arXiv preprint arXiv:2511.11407},
  year={2025}
}

@article{singhal2023large,
  title={Large language models encode clinical knowledge},
  author={Singhal, Karan and Azizi, Shekoofeh and Tu, Tao and Mahdavi, S Sara and Wei, Jason and Chung, Hyung Won and Scales, Nathan and Tanwani, Ajay and Cole-Lewis, Heather and Pfohl, Stephen and others},
  journal={Nature},
  volume={620},
  number={7972},
  pages={172--180},
  year={2023},
  publisher={Nature Publishing Group UK London}
}

@inproceedings{huang2024visual,
  title={Visual hallucinations of multi-modal large language models},
  author={Huang, Wen and Liu, Hongbin and Guo, Minxin and Gong, Neil},
  booktitle={Findings of the Association for Computational Linguistics: ACL 2024},
  pages={9614--9631},
  year={2024}
}

@inproceedings{soni2022radqa,
  title={Radqa: A question answering dataset to improve comprehension of radiology reports},
  author={Soni, Sarvesh and Gudala, Meghana and Pajouhi, Atieh and Roberts, Kirk},
  booktitle={Proceedings of the thirteenth language resources and evaluation conference},
  pages={6250--6259},
  year={2022}
}

@inproceedings{ben2019vqa,
  title={Vqa-med: Overview of the medical visual question answering task at imageclef 2019},
  author={Ben Abacha, Asma and Hasan, Sadid A and Datla, Vivek V and Demner-Fushman, Dina and M{\"u}ller, Henning},
  booktitle={Proceedings of CLEF (Conference and Labs of the Evaluation Forum) 2019 Working Notes},
  year={2019},
  organization={9-12 September 2019}
}

@article{zhang2023biomedclip,
  title={Biomedclip: a multimodal biomedical foundation model pretrained from fifteen million scientific image-text pairs},
  author={Zhang, Sheng and Xu, Yanbo and Usuyama, Naoto and Xu, Hanwen and Bagga, Jaspreet and Tinn, Robert and Preston, Sam and Rao, Rajesh and Wei, Mu and Valluri, Naveen and others},
  journal={arXiv preprint arXiv:2303.00915},
  year={2023}
}

@article{nori2023can,
  title={Can generalist foundation models outcompete special-purpose tuning? case study in medicine},
  author={Nori, Harsha and Lee, Yin Tat and Zhang, Sheng and Carignan, Dean and Edgar, Richard and Fusi, Nicolo and King, Nicholas and Larson, Jonathan and Li, Yuanzhi and Liu, Weishung and others},
  journal={arXiv preprint arXiv:2311.16452},
  year={2023}
}

@article{cole2019brain,
  title={Brain age and other bodily ‘ages’: implications for neuropsychiatry},
  author={Cole, James H and Marioni, Riccardo E and Harris, Sarah E and Deary, Ian J},
  journal={Molecular psychiatry},
  volume={24},
  number={2},
  pages={266--281},
  year={2019},
  publisher={Nature Publishing Group UK London}
}

@article{jack1999prediction,
  title={Prediction of AD with MRI-based hippocampal volume in mild cognitive impairment},
  author={Jack Jr, Clifford R and Petersen, Ronald C and Xu, Yue Cheng and O’Brien, Peter C and Smith, Glenn E and Ivnik, Robert J and Boeve, Bradley F and Waring, Stephen C and Tangalos, Eric G and Kokmen, Emre},
  journal={Neurology},
  volume={52},
  number={7},
  pages={1397--1397},
  year={1999},
  publisher={Lippincott Williams \& Wilkins}
}

@article{habes2016white,
  title={White matter hyperintensities and imaging patterns of brain ageing in the general population},
  author={Habes, Mohamad and Erus, Guray and Toledo, Jon B and Zhang, Tianhao and Bryan, Nick and Launer, Lenore J and Rosseel, Yves and Janowitz, Deborah and Doshi, Jimit and Van der Auwera, Sandra and others},
  journal={Brain},
  volume={139},
  number={4},
  pages={1164--1179},
  year={2016},
  publisher={Oxford University Press}
}

@article{kong2018mapping,
  title={Mapping cortical brain asymmetry in 17,141 healthy individuals worldwide via the ENIGMA Consortium},
  author={Kong, Xiang-Zhen and Mathias, Samuel R and Guadalupe, Tulio and ENIGMA Laterality Working Group and Glahn, David C and Franke, Barbara and Crivello, Fabrice and Tzourio-Mazoyer, Nathalie and Fisher, Simon E and Thompson, Paul M and others},
  journal={Proceedings of the National Academy of Sciences},
  volume={115},
  number={22},
  pages={E5154--E5163},
  year={2018},
  publisher={National Academy of Sciences}
}

@article{yuksel2018longitudinal,
  title={Longitudinal brain volume changes in major depressive disorder},
  author={Y{\"u}ksel, Dilara and Engelen, Jennifer and Schuster, Verena and Dietsche, Bruno and Konrad, Carsten and Jansen, Andreas and Dannlowski, Udo and Kircher, Tilo and Krug, Axel},
  journal={Journal of Neural Transmission},
  volume={125},
  number={10},
  pages={1433--1447},
  year={2018},
  publisher={Springer}
}

@article{liu2023visual,
  title={Visual instruction tuning},
  author={Liu, Haotian and Li, Chunyuan and Wu, Qingyang and Lee, Yong Jae},
  journal={Advances in neural information processing systems},
  volume={36},
  pages={34892--34916},
  year={2023}
}

@article{alayrac2022flamingo,
  title={Flamingo: a visual language model for few-shot learning},
  author={Alayrac, Jean-Baptiste and Donahue, Jeff and Luc, Pauline and Miech, Antoine and Barr, Iain and Hasson, Yana and Lenc, Karel and Mensch, Arthur and Millican, Katherine and Reynolds, Malcolm and others},
  journal={Advances in neural information processing systems},
  volume={35},
  pages={23716--23736},
  year={2022}
}

@inproceedings{radford2021learning,
  title={Learning transferable visual models from natural language supervision},
  author={Radford, Alec and Kim, Jong Wook and Hallacy, Chris and Ramesh, Aditya and Goh, Gabriel and Agarwal, Sandhini and Sastry, Girish and Askell, Amanda and Mishkin, Pamela and Clark, Jack and others},
  booktitle={International conference on machine learning},
  pages={8748--8763},
  year={2021},
  organization={PmLR}
}

@article{team2023gemini,
  title={Gemini: a family of highly capable multimodal models},
  author={Team, Gemini and Anil, Rohan and Borgeaud, Sebastian and Alayrac, Jean-Baptiste and Yu, Jiahui and Soricut, Radu and Schalkwyk, Johan and Dai, Andrew M and Hauth, Anja and Millican, Katie and others},
  journal={arXiv preprint arXiv:2312.11805},
  year={2023}
}

@article{claude3,
  title={The Claude 3 model family: Opus, Sonnet, Haiku},
  author={Anthropic},
  journal={Technical Report},
  year={2024}
}

@article{cohen1960coefficient,
  title={A coefficient of agreement for nominal scales},
  author={Cohen, Jacob},
  journal={Educational and psychological measurement},
  volume={20},
  number={1},
  pages={37--46},
  year={1960},
  publisher={Sage Publications Sage CA: Thousand Oaks, CA}
}

@article{vepa2025multimodal,
  title={A Multimodal LLM Approach for Visual Question Answering on Multiparametric 3D Brain MRI},
  author={Vepa, Arvind Murari and Yu, Yannan and Gan, Jingru and Cuturrufo, Anthony and Li, Weikai and Wang, Wei and Scalzo, Fabien and Sun, Yizhou},
  journal={arXiv preprint arXiv:2509.25889},
  year={2025}
}

@article{guo2026mm,
  title={MM-NeuroOnco: A Multimodal Benchmark and Instruction Dataset for MRI-Based Brain Tumor Diagnosis},
  author={Guo, Feng and Liu, Jiaxiang and Li, Yang and Shi, Qianqian and Xu, Mingkun},
  journal={arXiv preprint arXiv:2602.22955},
  year={2026}
}

@article{gai20253d,
  title={3d-rad: A comprehensive 3d radiology med-vqa dataset with multi-temporal analysis and diverse diagnostic tasks},
  author={Gai, Xiaotang and Liu, Jiaxiang and Li, Yichen and Meng, Zijie and Wu, Jian and Liu, Zuozhu},
  journal={arXiv preprint arXiv:2506.11147},
  year={2025}
}

@article{jiang2026lumen,
  title={LUMEN: Longitudinal Multi-Modal Radiology Model for Prognosis and Diagnosis},
  author={Jiang, Zhifan and Yang, Dong and Nath, Vishwesh and Parida, Abhijeet and Kulkarni, Nishad P and Xu, Ziyue and Xu, Daguang and Anwar, Syed Muhammad and Roth, Holger R and Linguraru, Marius George},
  journal={arXiv preprint arXiv:2602.21142},
  year={2026}
}

@article{wang2024enhancing,
  title={Enhancing vision-language models for medical imaging: bridging the 3D gap with innovative slice selection},
  author={Wang, Yuli and Peng, Jian and Dai, Yuwei and Jones, Craig and Sair, Haris and Shen, Jinglai and Loizou, Nicolas and Wu, Jing and Hsu, Wen-Chi and Imami, Maliha and others},
  journal={Advances in Neural Information Processing Systems},
  volume={37},
  pages={99947--99964},
  year={2024}
}

@article{lei2024autorg,
  title={Autorg-brain: Grounded report generation for brain mri},
  author={Lei, Jiayu and Zhang, Xiaoman and Wu, Chaoyi and Dai, Lisong and Zhang, Ya and Zhang, Yanyong and Wang, Yanfeng and Xie, Weidi and Li, Yuehua},
  journal={arXiv preprint arXiv:2407.16684},
  year={2024}
}

@article{yip2025medbookvqa,
  title={MedBookVQA: A Systematic and Comprehensive Medical Benchmark Derived from Open-Access Book},
  author={Yip, Sau Lai and He, Sunan and Nie, Yuxiang and Chan, Shu Pui and Ye, Yilin and Lam, Sum Ying and Chen, Hao},
  journal={arXiv preprint arXiv:2506.00855},
  year={2025}
}

@article{bai2024m3d,
  title={M3d: Advancing 3d medical image analysis with multi-modal large language models},
  author={Bai, Fan and Du, Yuxin and Huang, Tiejun and Meng, Max Q-H and Zhao, Bo},
  journal={arXiv preprint arXiv:2404.00578},
  year={2024}
}

@inproceedings{yue2024mmmu,
  title={Mmmu: A massive multi-discipline multimodal understanding and reasoning benchmark for expert agi},
  author={Yue, Xiang and Ni, Yuansheng and Zhang, Kai and Zheng, Tianyu and Liu, Ruoqi and Zhang, Ge and Stevens, Samuel and Jiang, Dongfu and Ren, Weiming and Sun, Yuxuan and others},
  booktitle={Proceedings of the IEEE/CVF conference on computer vision and pattern recognition},
  pages={9556--9567},
  year={2024}
}

@misc{abbasi2025smri,
  author       = {Mohammad H. Abbasi and Ehsan Adeli},
  title = {s{MRI} {P}rocessing {P}ipeline: A lightweight, end-to-end workflow for structural brain {MRI} preprocessing and quality control},
  year         = {2025},
  institution  = {Stanford Translational AI Lab, Stanford University},
  publisher    = {Zenodo},
  doi          = {10.5281/zenodo.17503175},
  url          = {https://doi.org/10.5281/zenodo.17503175},
  note = {Zenodo, doi: 10.5281/zenodo.17503175},
 }

@misc{nerrise2026geosae,
      title={GeoSAE: Geometric Prior-Guided Layer-Wise Sparse Autoencoder Annotation of Brain MRI Foundation Models}, 
      author={Favour Nerrise and Lucy Yin and Mohammad H. Abbasi and Kilian M. Pohl and Ehsan Adeli},
      year={2026},
      eprint={2605.01829},
      archivePrefix={arXiv},
      primaryClass={cs.CV},
      url={https://arxiv.org/abs/2605.01829}, 
}

@article{endo2024data,
  title={Data-driven discovery of movement-linked heterogeneity in neurodegenerative diseases},
  author={Endo, Mark and Nerrise, Favour and Zhao, Qingyu and Sullivan, Edith V and Fei-Fei, Li and Henderson, Victor W and Pohl, Kilian M and Poston, Kathleen L and Adeli, Ehsan},
  journal={Nature machine intelligence},
  volume={6},
  number={9},
  pages={1034--1045},
  year={2024},
  publisher={Nature Publishing Group UK London}
}

@article{landis1977measurement,
  title={The measurement of observer agreement for categorical data},
  author={Landis, J Richard and Koch, Gary G},
  journal={biometrics},
  pages={159--174},
  year={1977},
  publisher={JSTOR}
}
}
\clearpage

\clearpage

\clearpage
\appendix
\etocdepthtag.toc{supplement}

\etocsettagdepth{mainmatter}{none}
\etocsettagdepth{supplement}{subsection}
\etocsettocstyle
  {\begin{center}\large\bfseries Contents of Appendix/Supplementary Material\end{center}\vspace{0.5em}}
  {}
\tableofcontents
\clearpage

\section{Limitations and Future Work}
\label{app:limitations}

\myparagraph{Scope limits.} Four scope limits bound the claims of this benchmark. First, white matter hyperintensity items use T1-weighted volumes while T2 or FLAIR are clinically preferred, so WMH items carry \texttt{answerable\_modalities = T2, FLAIR} and can be scored separately when additional modalities are available. Second, some Measurement items require volumetric precision beyond human visual capability, are tagged \texttt{human\_assessable = false}, and are intentionally retained to measure capabilities a clinician cannot. Third, shortcut refinement drives text-only accuracy to within 5.1 points of random, and the residual gap reflects irreducible biological priors (rarity of severe WMH in healthy subjects, mild structural asymmetry) that cannot be removed without discarding clinically meaningful items. Fourth, the human visual reference rests on two clinicians (a radiology resident and a neurosurgery resident) reading 100 items, which yields fair-to-moderate inter-rater agreement ($\kappa = 0.40$ per~\cite{landis1977measurement}) and bounds the claim to a perceptual ceiling under resident-level reading rather than a board-certified attending-level estimate; the per-category structure (Table~\ref{tab:clinician}) was independently consistent across both raters, but a larger and more senior rater pool would tighten the bound. Tier~2 reproducibility further depends on users running a FreeSurfer preprocessing pipeline compatible with the shipped labels, and \projectname{} scores recognition and structural reasoning rather than treatment planning, prognostic inference, or report generation.

\myparagraph{Future work.} Several extensions follow naturally from the current scope. (i) Multimodal expansion of the WMH category to T2 and FLAIR sequences, paired with a comparison of T1-only vs.\ multimodal VLM scoring on the same subjects, would convert the modality limit from a constraint into a controlled axis of evaluation. (ii) An attending-level rater study, with $\geq 5$ board-certified radiologists or neuroradiologists scoring an expanded 250-item subset, would sharpen the human visual ceiling and allow a direct comparison between trainee and attending performance per category. (iii) A longitudinal-only sub-track that scores VLMs on multi-timepoint reasoning (currently 1{,}166 items, $\sim2\%$ of the benchmark) is a natural extension as more longitudinal datasets enter the pool. (iv) Beyond recognition, downstream tasks such as report generation and treatment-plan reasoning are deferred to future benchmarks because they require ground truth that \projectname{}'s deterministic pipeline cannot supply without introducing reader-level phrasing drift.

\section{Complete Quality Rule List}
\label{app:rules}

\begin{table}[h]
\centering
\small
\caption{All 38 quality rules applied by the deterministic cleanup pipeline, organized by development phase. Rule IDs are non-contiguous because R18, R19, R21, R22, and R26 were drafted during pipeline iteration but later subsumed by R23--R25 (shortcut balancing) or R36 (path/ID validation) before the final freeze; the IDs are retained to preserve traceability against earlier internal audit logs.}
\begin{tabular}{lp{8.5cm}r}
\toprule
Rule & Description & Items \\
\midrule
\multicolumn{3}{l}{\textit{Expert-driven quality (Phases 1--2)}} \\
R1 & BraTS laterality convention alignment & 0 \\
R2/R3 & Low-asymmetry laterality removal ($<$10\% difference) & 1,794 \\
R4 & Cerebellar to posterior fossa terminology & 21 \\
R4b/R4c & Report negation detection and signal extraction & 254 \\
R5--R6 & Multi-lobe parsing, L/R answer removal & 27 \\
R7--R9 & Clinical wording, age context, longitudinal context & 8,781 \\
R10 & Longitudinal direction verification against volumes & 1,138 \\
R11--R12 & Compound MCQ options; dataset name removal & 604 \\
R13 & Mislabeled BraTS-MEN subject removal & 30 \\
R14 & NaN FreeSurfer volume removal & 591 \\
R15 & Session alignment (latest session with image + FreeSurfer) & 52,035 \\
\midrule
\multicolumn{3}{l}{\textit{Machine-verified quality (7 audit rounds)}} \\
R27 & Multi-session volume/age/signal fix & 586 \\
R28b & Per-dataset age range label normalization & 1,182 \\
R29--R30 & Measurement boundary fix; duplicate answer removal & 46 \\
R31 & Open-ended Yes/No to YN type conversion & 2,448 \\
R32 & MCQ ``None'' to ``No changes'' label clarity & 382 \\
R33 & WholeBrain ratio removal (HCP-Aging, HCP-Dev) & 337 \\
R34/R38 & Age answer direct fix (ABCD, ADNI, AIBL) & 37 \\
R35 & BraTS signal MCQ from parsed report features & 124 \\
R36 & Image path validation, subject ID types, IXI modality & 21,322 \\
R37 & Longitudinal session ordering and answer verification & 2,805 \\
\midrule
\multicolumn{3}{l}{\textit{Shortcut elimination}} \\
R23 & YN 50/50 balance per template per dataset & 7,725 \\
R24 & MCQ/Open shortcut removal & 4,485 \\
R25 & MCQ position balance (25\% per position per dataset) & All MCQ \\
R28 & MCQ standardization (all items have exactly 4 options) & 1,263 \\
\midrule
\multicolumn{3}{l}{\textit{Annotations}} \\
R16 & Answerable modalities field per question & All \\
R17 & Human-assessable flag per question & All \\
R20 & Neurological convention (RAS) verification & All \\
\bottomrule
\end{tabular}
\end{table}

\section{Dataset Composition and Splits}
\label{app:dataset}

\begin{figure}[h]
    \centering
    \includegraphics[width=\textwidth]{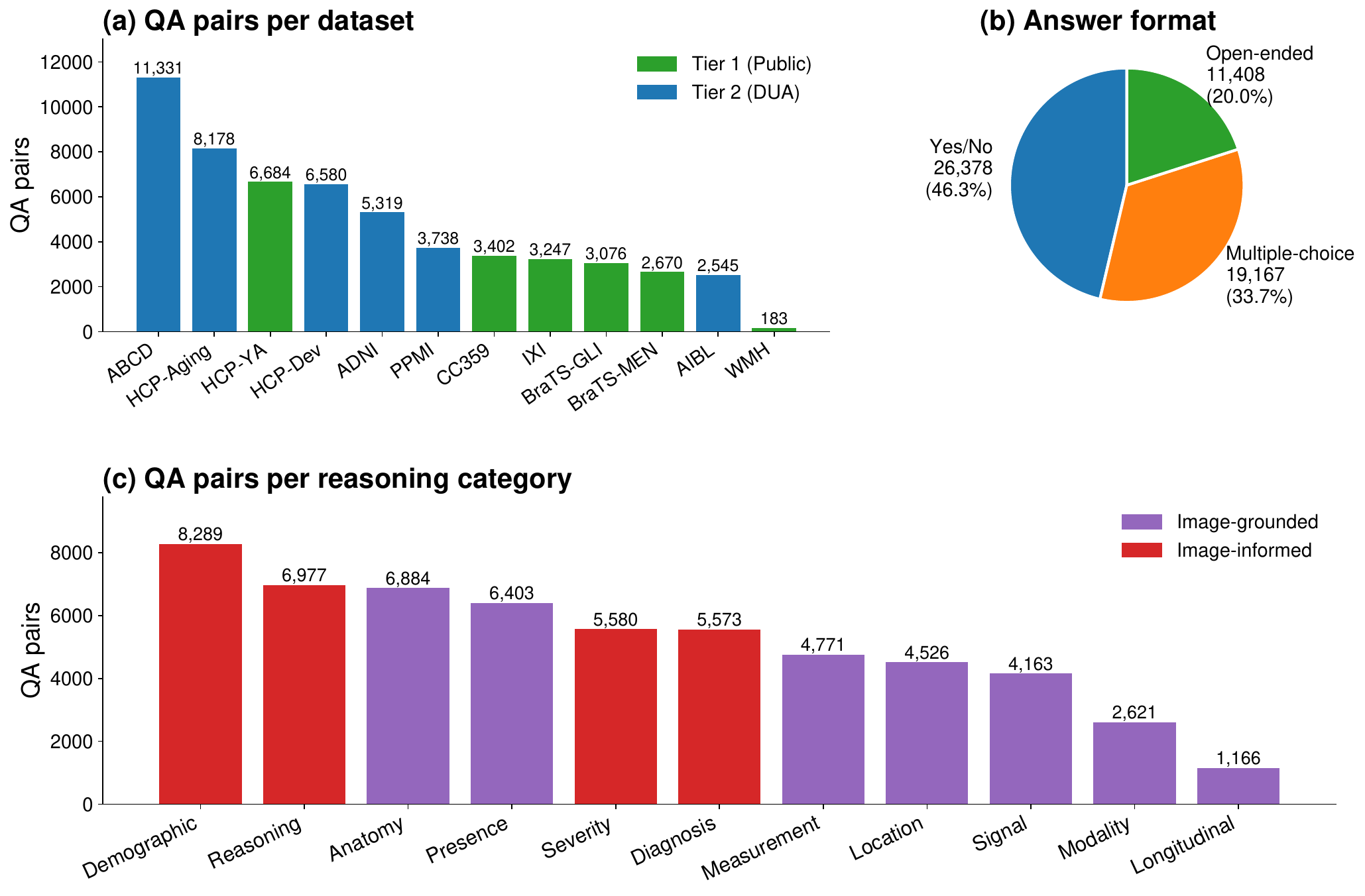}
    \caption{\projectname{} benchmark statistics. (a) QA pairs per dataset across the 12 source datasets, colored by release tier: Tier~1 Public (green) and Tier~2 DUA-restricted (blue). (b) QA pairs per reasoning category across 11 categories, colored by image dependency: \ig\ (answerable from visual inspection) and \ii\ (requires quantitative volumetry or clinical instruments). (c) Answer format distribution across 56{,}953 items: Yes/No (46.3\%), four-option MCQ (33.7\%), and open-ended (20.0\%). (d) Scale and breadth comparison against existing medical VQA benchmarks.}
    \label{fig:dataset_stats}
\end{figure}

Table~\ref{tab:per_dataset} reports the per-dataset test-public item counts and release tier; Figure~\ref{fig:dataset_stats} shows the per-dataset, per-category, and per-format breakdown alongside a comparison with existing benchmarks. No subject appears in more than one split.

\begin{table}[h]
\caption{Per-dataset composition of \projectname{}. Full pool shows all 56{,}953 items; test-public shows the 5{,}524 evaluation items. Text-only is the majority-per-template floor on the full pool.}
\label{tab:per_dataset}
\centering
\small
\begin{tabular}{llrrc}
\toprule
Dataset & Tier & Full Pool & Test-Public & Text-only \\
\midrule
ABCD & DUA & 11{,}331 & 1{,}112 & 50.5\% \\
ADNI & DUA & 5{,}319 & 511 & 50.7\% \\
AIBL & DUA & 2{,}545 & 293 & 49.1\% \\
BraTS-GLI & Pub & 3{,}076 & 294 & 43.6\% \\
BraTS-MEN & Pub & 2{,}670 & 254 & 45.6\% \\
CC359 & Pub & 3{,}402 & 323 & 47.3\% \\
HCP-Aging & DUA & 8{,}178 & 791 & 49.4\% \\
HCP-Dev & DUA & 6{,}580 & 678 & 48.0\% \\
HCP-YA & Pub & 6{,}684 & 614 & 50.0\% \\
IXI & Pub & 3{,}247 & 308 & 45.5\% \\
PPMI & DUA & 3{,}738 & 326 & 51.3\% \\
WMH & Pub & 183 & 20 & 60.1\% \\
\midrule
\textbf{Overall} & & \textbf{56{,}953} & \textbf{5{,}524} & \textbf{44.6\%} \\
\bottomrule
\end{tabular}
\end{table}

\section{Leaderboard and Public Release}
\label{app:leaderboard}

The leaderboard at \url{https://neuroqa.stanford.edu} accepts submissions on the test-private split (5,581 items with hidden ground truth) and reports per-dataset, per-category, and per-format accuracy. Tier~1 QA pairs, generation scripts, the evaluation toolkit, and the clinician evaluation interface are released at the same URL; Tier~2 users reproduce the full benchmark by obtaining their own data access and running the deterministic pipeline.

\myparagraph{Contamination handling.} Because Tier~2 reproducers regenerate QA pairs from the same source datasets, test-private subject IDs are in principle derivable from the underlying neuroimaging cohorts. The release policy therefore (i) publishes a hashed manifest of test-private subject IDs so that Tier~2 reproducers can exclude them at generation time, (ii) accepts submissions through a server-side evaluation endpoint that consumes only model predictions for published item IDs and never exposes ground-truth answers, and (iii) rate-limits submissions per team to discourage hill-climbing on the hidden split. Concrete mechanisms (hash format, endpoint specification, rate limit) are documented in the release README and versioned alongside each benchmark release.

\section{Expert Review Details}
\label{app:expert}

Two rounds of expert review established quality rules through systematic evaluation of benchmark items by domain experts. Below we detail the findings from each review phase.

\subsection{First Round: Template Review (3 experts, 24 items)}

Three domain experts independently reviewed 24 stratified items. Their feedback produced the following corrections:

\begin{figure}[h]
    \centering
    \includegraphics[width=\textwidth]{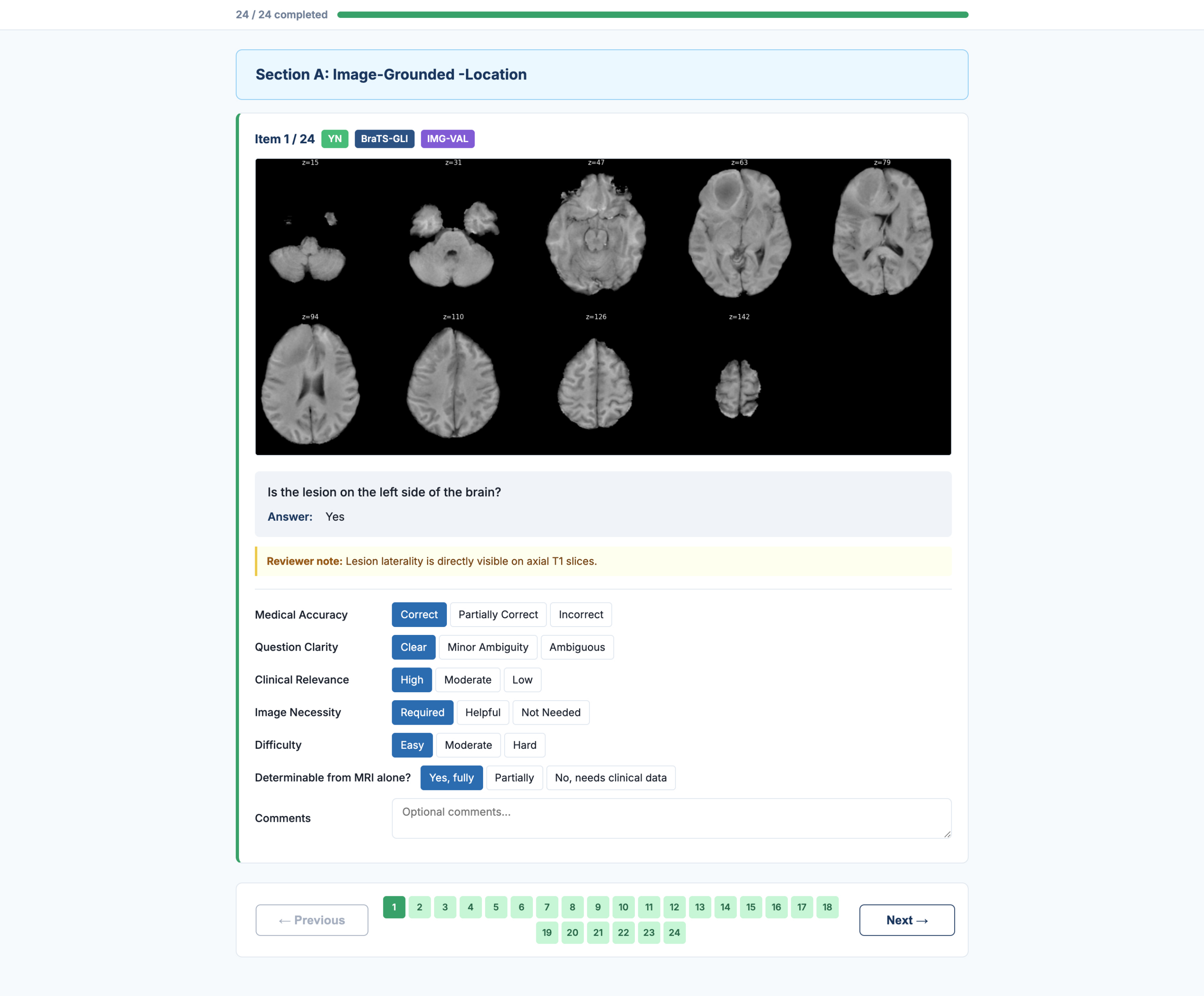}
    \caption{Phase 1 expert review interface. Three domain experts independently reviewed 24 stratified items across datasets and categories, rating accuracy, clarity, relevance, and clinical validity.}
    \label{fig:phase1}
\end{figure}

\paragraph{1. Laterality removal (BraTS-GLI/MEN, 786 QA removed).} Expert review flagged laterality labels as unreliable. Systematic verification of 220 unilateral BraTS-GLI subjects showed only 49\% agreement between report-stated and voxel-based laterality, effectively random, because BraTS reports and NIfTI images may originate from different processing stages or conventions. Lobe-identification questions were retained.

\paragraph{2. Cerebellopontine angle fix (BraTS-MEN, 28 QA corrected).} Fifteen BraTS-MEN subjects with cerebellopontine angle (CPA) lesions were incorrectly labeled as ``cerebellar'' in Location questions. CPA lesions are extra-axial posterior fossa masses, anatomically distinct from cerebellar parenchymal lesions. Corrected to ``posterior fossa.''

\paragraph{3. Subtle asymmetry removal (1,524 QA removed).} 57\% of laterality questions had less than 10\% volume asymmetry between left and right structures, differences imperceptible on visual inspection. Removed across 7 datasets; retained questions with clearly visible asymmetry ($>$10\% difference).

\paragraph{4. Signal wording standardization (654 QA reworded).} The neuroradiologist recommended standard clinical phrasing: ``What is the T1 signal intensity of the lesion?'' was revised to ``What is the signal intensity of the lesion on T1-weighted imaging?'' No answers changed; only question text was reworded to match radiology reporting conventions.

\paragraph{5--6. White matter on T1 and modality annotation.} Both reviewers noted that white matter assessment is conventionally performed on T2/FLAIR, not T1. Rather than removing 5,680 WMH signal questions, we added a \texttt{recommended\_modality} field. This allows users with T2 data to provide multi-modality input while preserving the benchmark for T1-only evaluation. For T1-only datasets, this is acknowledged as a limitation.

\paragraph{7--9. Measurement questions beyond visual inspection.} Both reviewers noted that volume estimation is ``impossible to judge visually.'' We intentionally retain Measurement questions (3,767 QA) because \projectname{} evaluates VLM capabilities beyond human visual inspection. Models operating on 3D volumetric data should extract quantitative information that clinicians cannot assess by eye. This design choice is documented as a deliberate scope decision.

\paragraph{10. Anatomy questions.} Both reviewers rated Anatomy questions as correct and highly relevant. The neuroradiologist noted that axial views are atypical for hippocampal assessment; however, this applies only to the 2D survey visualization. \projectname{} provides full 3D volumetric input to models, including all orientations.

\paragraph{11. Age context (6,445 QA updated).} Both reviewers noted that age is essential for interpreting structural brain questions. Added subject age to questions across ABCD, HCP-YA, HCP-Aging, and HCP-Dev. Example: ``Is the hippocampal volume below the expected range for this age group?'' became ``...for this 55-year-old subject?''

\paragraph{12. Clinical history limitation.} The neuroradiologist noted that clinical history (e.g., corticosteroid use) would aid interpretation. This metadata is unavailable in the datasets and is acknowledged as a limitation inherent to retrospective imaging benchmarks.

\paragraph{13. Compound MCQ options for Presence (872 QA updated).} Both reviewers noted that single-answer Presence MCQs are problematic when multiple findings coexist. In ADNI, 1,357 subjects have all three findings (atrophy + enlargement + WMH). Updated to include compound options based on each subject's actual FreeSurfer findings.

\paragraph{14--16. Longitudinal fixes (2,024 QA updated).} Both reviewers noted longitudinal questions need time context. The neuroradiologist flagged ambiguous phrasing (``remained stable or decreased,'' inclusive disjunction). Applied three fixes: time interval from metadata, subject age added, and 557 inclusive disjunctions corrected.

\paragraph{17. Mislabeled BraTS-MEN (22 QA removed).} The neuroradiologist flagged a BraTS-MEN subject as potentially mislabeled. Investigation confirmed 2 subjects with non-meningioma disease labels. All 22 QA pairs removed.

\paragraph{18. Diagnosis framework (392 QA updated).} Both reviewers noted that mixing MCI, AD, and FTD without specifying the diagnostic framework is ambiguous. Added explicit dataset-specific references to Diagnosis MCQ questions.

\subsection{Second Round: Clinician Review with Images (2 clinicians, 57 items)}

A radiology resident and a neuroradiology professor evaluated 57 items using a 3-plane NIfTI viewer. Items were selected to maximize coverage of \ig\ categories (Table~\ref{tab:grounded}).

\begin{table}[h]
\caption{Image dependency classification of 203 question templates. \ig[Image-grounded] categories are answerable from a 3-plane viewer; \ii\ categories require quantitative volumetry or clinical instruments. Two templates appear under two categories each (cohort-dependent evidence type), so per-category counts in Tables~\ref{tab:tmpl_anatomy}--\ref{tab:tmpl_signal} sum to 205.}

\label{tab:grounded}
\centering
\small
\begin{tabular}{lrrr}
\toprule
Category & Grounded & Informed & \% Grounded \\
\midrule
Modality & 2 & 0 & 100\% \\
Anatomy & 37 & 0 & 100\% \\
Location & 28 & 0 & 100\% \\
Measurement & 11 & 0 & 100\% \\
Signal & 12 & 0 & 100\% \\
Longitudinal & 14 & 0 & 100\% \\
Presence & 27 & 0 & 100\% \\
\midrule
Reasoning & 0 & 25 & 0\% \\
Diagnosis & 0 & 18 & 0\% \\
Severity & 0 & 19 & 0\% \\
Demographic & 0 & 10 & 0\% \\
\midrule
\textbf{Total} & \textbf{131} & \textbf{72} & \textbf{65\%} \\
\bottomrule
\end{tabular}
\end{table}

\begin{figure}[h]
    \centering
    \includegraphics[width=\textwidth]{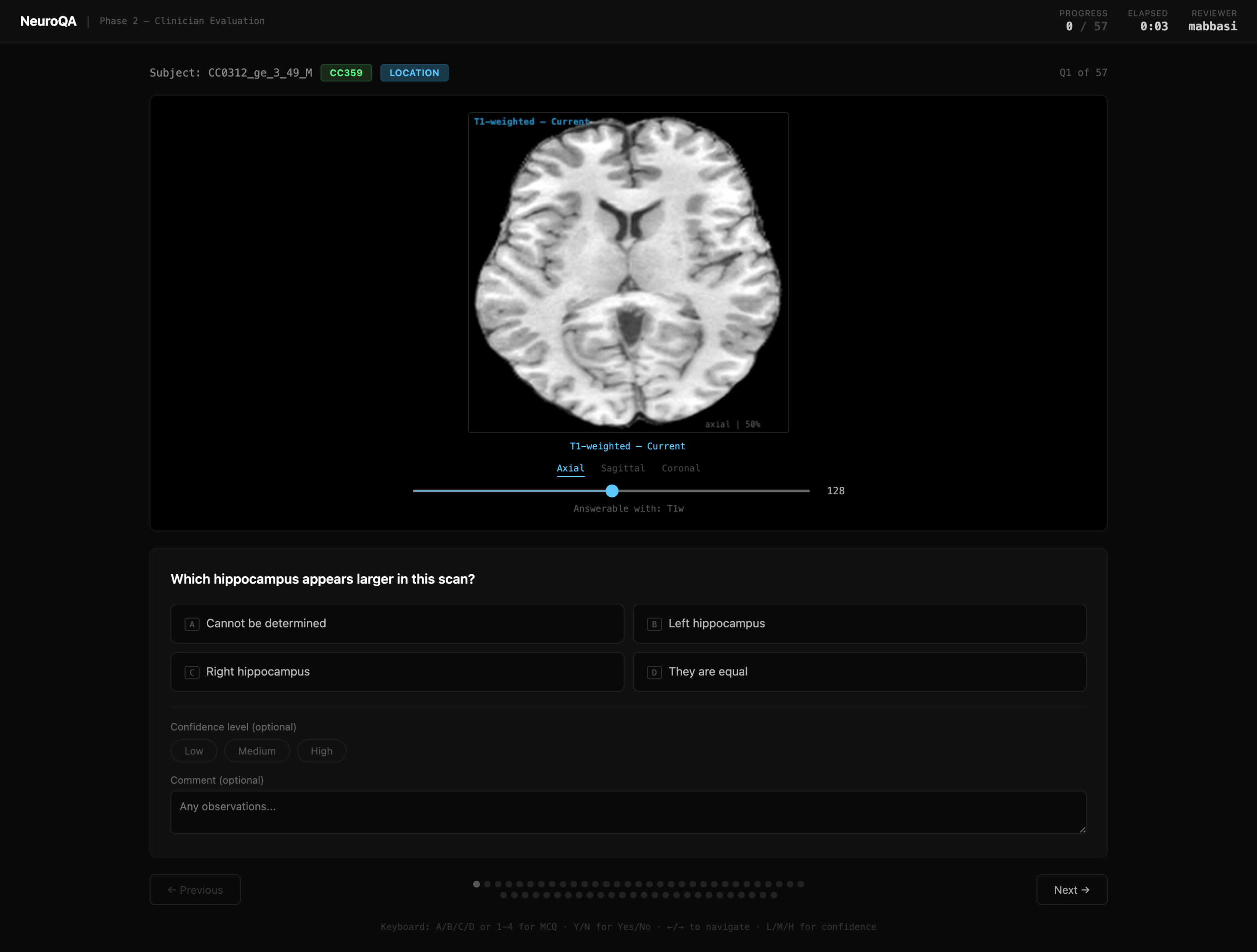}
    \caption{Phase 2 clinician review interface with 3-plane NIfTI brain viewer. Clinicians view axial, sagittal, and coronal planes with a slice navigator, then answer each question and rate confidence.}
    \label{fig:phase2}
\end{figure}

Phase 2 items were sampled to prioritize \ig\ categories: Location (15), Anatomy (10), Measurement (6), Presence (8), Longitudinal (6), Signal (5), Modality (2), and Diagnosis (5, BraTS tumor differential only). Categories where answers derive from clinical scores rather than imaging (Demographic, Severity) were excluded from the clinician review, as were primarily \ii\ categories (Reasoning). This yielded 57 items with an estimated 30--40 minutes per clinician.

This review identified two additional quality issues:

\paragraph{Report negation detection (R4b).} The parser incorrectly extracted positive findings from negated report phrases (e.g., ``no midline shift'' was parsed as midline shift present). Corrected the negation detection logic.

\paragraph{Signal extraction from reports (R4c).} Signal intensity was extracted from truncated CSV fields rather than the full Modal-wise Findings text. Extended extraction to use the complete report text, correcting 248 items.

\section{Clinician Evaluation Protocol}
\label{app:clinician}

\begin{figure}[h]
    \centering
   \includegraphics[width=\textwidth]{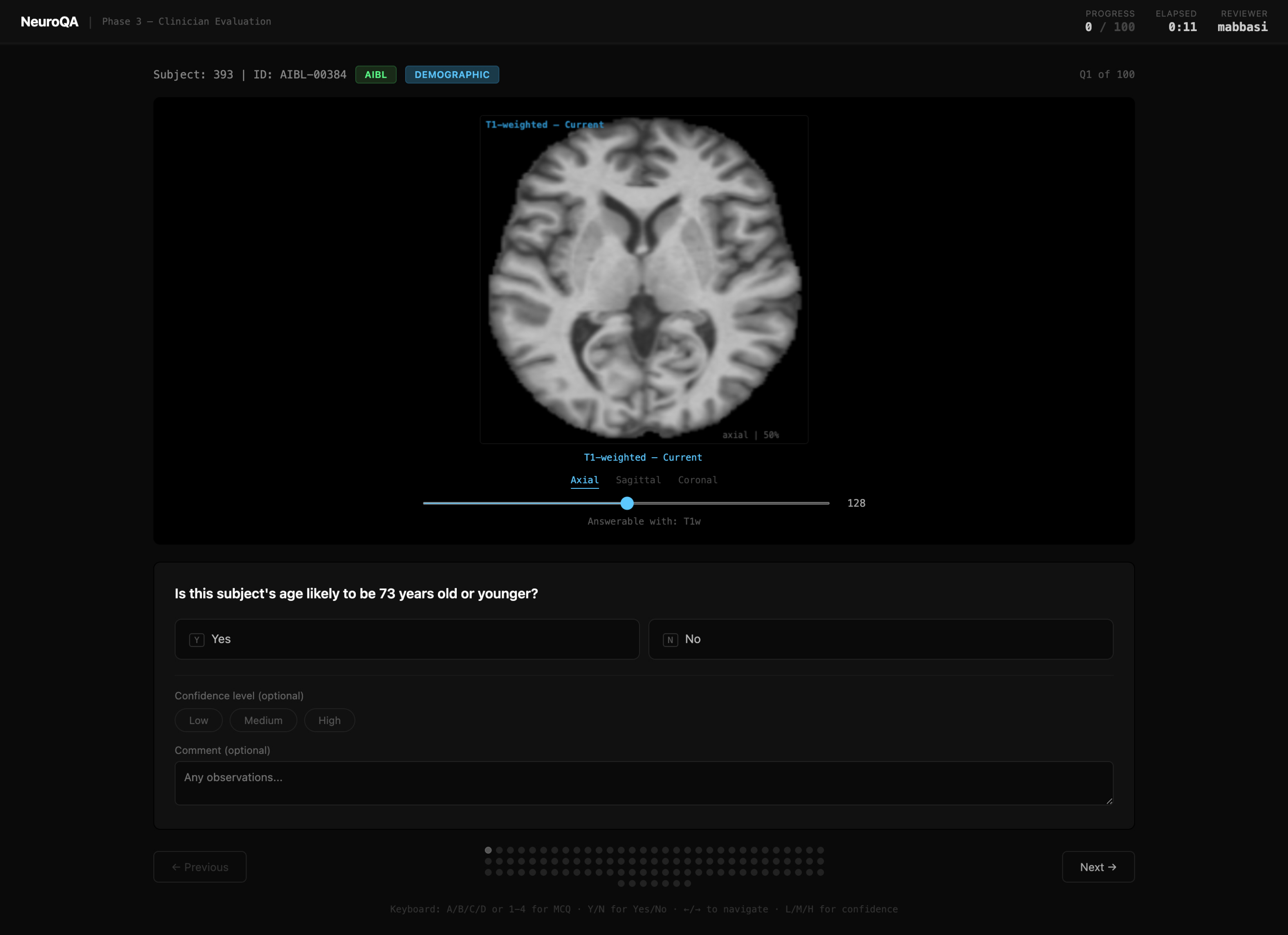}
    \caption{\projectname{} clinician evaluation interface. The 3-plane NIfTI viewer displays brain MRI in axial, sagittal, and coronal planes with a slice navigator. Clinicians answer each question independently. For multi-modality questions, both T1 and T2 scans are shown. For longitudinal questions, current and prior scans are displayed with dates.}
    \label{fig:survey}
\end{figure}

The Phase 3 study (\S\ref{sec:clinician_study}) sampled 100 items from 100 distinct question templates out of 178 human-assessable templates, with proportional stratification across all 11 categories. Sampled subjects were disjoint from those used in the first and second rounds of expert review. Figure~\ref{fig:survey} shows the web-based viewer used by both raters.

\subsection{Per-Category Clinician Analysis}
\label{app:clinician_analysis}

This section expands on the per-category patterns in Table~\ref{tab:clinician}.

\myparagraph{Where clinicians excel.}
\textbf{Modality} (100\%/100\%): both raters identify T1 vs.\ T2 correctly on all items, validating the protocol for directly observable information.
\textbf{Anatomy} (55.6\%/61.1\%, 67\% agreement): strong on gross morphology; the 5.5-pp gap reflects threshold disagreements on borderline cases.
\textbf{Longitudinal} (57.1\%/57.1\%): identical performance suggests visual comparison of paired volumes is a stable task even for subtle changes.

\myparagraph{Where clinicians plateau.}
\textbf{Reasoning} (28.6\%/28.6\%): both raters score below MCQ chance (25\%). Ground truth derives from FreeSurfer cutoffs not reproducible by eye; identical scores on the same wrong answers indicate a systematic perceptual ceiling.
\textbf{Severity} (28.6\%/42.9\%): labels are anchored to clinical instruments (CDR, UPDRS) without one-to-one correspondence to morphology. Rater~2's higher score likely reflects instrument familiarity, not improved visual extraction.

\myparagraph{Where raters diverge most.}
\textbf{Location} (27.3\%/54.5\%, 36\% agreement): the largest gap among \ig\ categories, concentrated on lobar boundary cases requiring cross-plane spatial reasoning. Rater~2 (neurosurgery) outperforms Rater~1 (radiology), consistent with surgical mapping experience.
\textbf{Diagnosis} (45.5\%/72.7\%, 18\% agreement): the largest overall divergence. Labels encode multi-modal workup (clinical history, biomarkers) no visual read can recover; the gap reflects specialty bias.

\myparagraph{Implications.}
These patterns confirm three benchmark properties: (1) the \ig/\ii\ split holds empirically; (2) below-chance performance on \ii\ categories validates that ground truth is not visually recoverable; (3) the 48.9\% closed mean is a meaningful ceiling for visual inspection alone.

\section{Open-Ended Scorer Validation}
\label{app:judge}

This appendix validates that the automated open-ended scoring metric (token-level F1) ranks model outputs consistently with human judgment.

\myparagraph{Sample.} Open-ended items were scored on a stratified subset of $N{=}100$ items drawn from test-public, covering seven of the eight open-ended categories (all except Presence, which is not represented in the GPT-5-nano response release that anchors the sample).

\myparagraph{Models scored.} To validate the metric across the leaderboard's accuracy range rather than at a single point, raters scored model outputs from three VLMs spanning the open-EM distribution observed in Table~\ref{tab:open_metrics}: \textbf{GPT-5.2} (31.3\% open EM, top of range), \textbf{MedGemma-4B} (18.6\% open EM, mid-low range with clinical-domain writing register), and \textbf{GPT-5-nano} (16.0\% open EM, bottom of range). Each rater scored $100 \times 3 = 300$ (item, model) rows, of which $291$ are populated; $9$ rows (1 GPT-5.2, 8 MedGemma-4B) have no response in the public dump and are excluded from per-row statistics. All three models are scored in the stack vision condition (matching the leaderboard convention).

\myparagraph{Rubric.} For each (question, gold reference, model answer) row, the rater decides whether the model answer conveys the same information as the gold reference on the specific fact the question asks about. Verdict is binary: \textbf{1} (correct, including paraphrase or synonym) or \textbf{0} (incorrect, including refusals, off-topic responses, and contradictions of the gold). Wording differences, hedging, and unrelated additional content do not affect the verdict. The rater does not assess clinical correctness of the gold itself; the gold is treated as authoritative. Edge cases follow the protocol below:
\begin{itemize}[leftmargin=*, nosep]
    \item Partial matches (e.g., correct lobe but wrong side) are rated \textbf{0} with a free-text note, since binary judgments target sharp agreement; partial credit is reflected in the automated F1 score.
    \item Refusals or non-committal answers (``I cannot tell from the image'') are rated \textbf{0}.
    \item Synonym matches (gold ``left frontal'' vs. model ``frontal lobe on the left'') are rated \textbf{1}.
    \item Numeric ranges in the gold are rated \textbf{1} when the model emits any value within the range.
\end{itemize}

\begin{figure}[!t]
\centering
\begin{minipage}[t]{0.46\textwidth}
\centering
\includegraphics[width=\linewidth]{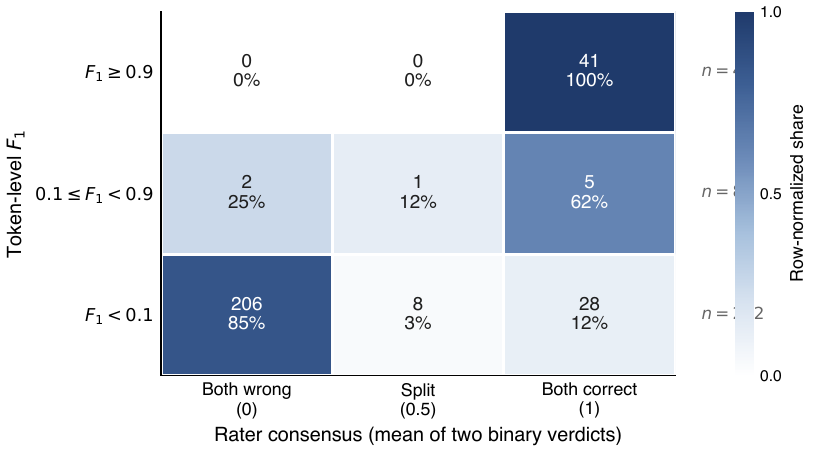}
\subcaption{F1-bucket by rater-consensus heatmap over the $291$ populated rows. Cell text is the row count and row-normalized proportion; cell color is the same proportion. The mass concentrates on the diagonal (F1\,${<}$\,0.1 with both raters incorrect; F1\,${\geq}$\,0.9 with both raters correct), and the off-diagonal $28$ rows in the F1\,${<}$\,0.1 row with consensus 1 are the paraphrase failures EM and F1 share.}
\label{fig:judge_scatter}
\end{minipage}\hfill
\begin{minipage}[t]{0.50\textwidth}
\centering
\includegraphics[width=\linewidth]{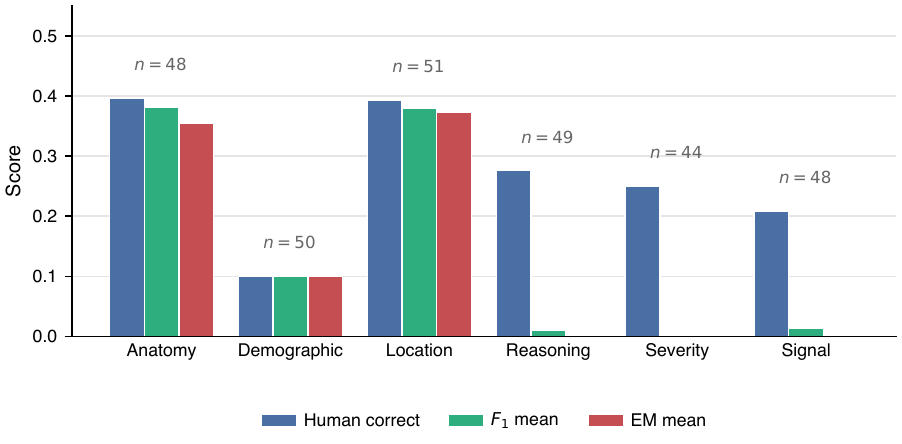}
\subcaption{Per-category mean rater correctness alongside automated F1 and EM means. Anatomy and Demographic align closely; Signal and Reasoning expose the F1/EM collapse on long descriptive gold answers.}
\label{fig:judge_per_category}
\end{minipage}
\caption{Open-ended scorer validation summary. Left: F1 vs.\ rater-consensus calibration as a $3{\times}3$ heatmap. Right: per-category alignment between human correctness and automated metrics.}
\label{fig:judge_summary}
\end{figure}

\myparagraph{Inter-rater agreement.} Cohen's $\kappa$ between the two raters was $\kappa{=}0.92$ with $96.9\%$ raw agreement, computed on all $n{=}291$ populated rows. Per-category $\kappa$ is reported in Table~\ref{tab:judge_breakdown}; the lower Severity value ($0.76$) tracks rater-judgment differences on the cognitive-level templates, where calls between Average, Below average, and Above median admit more leniency than the categorical Demographic and Anatomy items.

\myparagraph{Automated-metric correlation.} Spearman rank correlation between token-level F1 and the rater-consensus verdict was $\rho{=}0.74$ over the $291$ populated rows; the underlying calibration is shown as a $3{\times}3$ heatmap in Figure~\ref{fig:judge_summary}~(left). Exact match was weaker ($\rho{=}0.67$), with the gap concentrated on Signal items where EM cannot credit the paraphrase that F1 partially recovers. The result holds across the three model strata (GPT-5.2 $\rho{=}0.74$ on $n{=}99$, MedGemma-4B $\rho{=}0.87$ on $n{=}92$, GPT-5-nano $\rho{=}0.64$ on $n{=}100$). The MedGemma-4B value is highest because the model emits short stereotyped strings (``The brain'', ``Bright'', ``No''), so F1 and the human verdict collapse onto the same low values together. The GPT-5-nano value is the lowest because nano's short answers occasionally string-match a wrong gold (e.g., ``hypointense'' against an ``isointense'' gold), where F1 awards full credit while the rater scores zero.

\myparagraph{Per-category breakdown.}
Figure~\ref{fig:judge_summary}~(right) summarizes per-category alignment between rater correctness and the automated metrics; Table~\ref{tab:judge_breakdown} reports the underlying numbers and per-category Cohen's $\kappa$.

\begin{table}[h]
\caption{Open-ended scorer validation: per-category mean rater correctness, automated F1 and EM means, and Cohen's $\kappa$ on the populated rows. ``Human correct'' is the rater-consensus rate (mean of the two binary verdicts per row). $\kappa$ reads ``--'' for Longitudinal because only one row was rated by both judges.}
\label{tab:judge_breakdown}
\centering
\scriptsize
\setlength{\tabcolsep}{4pt}
\begin{tabular}{lrrrrr}
\toprule
Category & $n$ & Human correct & F1 mean & EM mean & Cohen's $\kappa$ \\
\midrule
Anatomy      & 48  & 0.396 & 0.382 & 0.354 & 1.00 \\
Demographic  & 50  & 0.100 & 0.100 & 0.100 & 1.00 \\
Location     & 51  & 0.392 & 0.380 & 0.373 & 0.92 \\
Longitudinal & 1   & 0.000 & 0.000 & 0.000 & --   \\
Reasoning    & 49  & 0.276 & 0.010 & 0.000 & 0.85 \\
Severity     & 44  & 0.250 & 0.000 & 0.000 & 0.76 \\
Signal       & 48  & 0.208 & 0.014 & 0.000 & 1.00 \\
\midrule
\textbf{All} & \textbf{291} & \textbf{0.275} & \textbf{0.151} & \textbf{0.142} & \textbf{0.92} \\
\bottomrule
\end{tabular}
\end{table}

The Signal gap visible in Figure~\ref{fig:judge_summary}~(right) drives our metric choice. F1 is rank-preserving within each stratum (hence the strong $\rho$) but its absolute scale collapses on long descriptive gold answers, which is why the leaderboard reports the closed-format aggregate as the headline metric.

\myparagraph{Limitations of the validation.}
The judges were non-clinician raters; their verdicts validate \emph{text-equivalence} between model output and gold reference but do not constitute an independent clinical assessment of either. Items where the gold itself is a graded clinical instrument (CDR, UPDRS) are scored under the same rubric without re-validating the gold. This bounds the claim to ``the open-ended scoring metric ranks model outputs the way a careful reader would rank them against the released gold,'' not ``the released gold is itself clinically optimal,'' which is established separately by \S\ref{sec:pipeline}.

\section{Supervised 3D CNN Baseline Details}
\label{app:cnn}

We provide implementation details for the multi-task 3D CNN reported in Table~\ref{tab:baselines}. The model is trained from scratch on the train split (40{,}355 items) and serves as a controlled supervised reference for what a small volumetric model recovers on \projectname{} without external pretraining.

\myparagraph{Architecture.} A custom 4-stage 3D CNN uses two $3{\times}3{\times}3$ convolutions, batch normalization, ReLU, and $2{\times}2{\times}2$ max-pool per stage. Channel widths progress $1 \to 16 \to 32 \to 64 \to 128$, followed by 3D adaptive average pooling to a 128-d feature. A 64-d template embedding, using the same digit-stripped template index as the text-only baseline, is concatenated to produce a 192-d shared representation. Three parallel heads (Linear $192{\to}128$, ReLU, Dropout 0.3, Linear $128{\to}d_t$) use $d_t \in \{2, 4, 379\}$ for YN, MCQ, and Open respectively. The Open head uses a closed vocabulary of 379 answers built from training data. Total parameters: \textbf{1.02M}.

\myparagraph{Training.} Each volume is resized to $96^3$ voxels via trilinear interpolation and min-max normalized to $[0,1]$. The model is trained with Adam (learning rate $10^{-4}$, no weight decay), batch size 8, for 15 epochs. Per-task cross-entropy losses are weighted by task frequency in each batch, and the best checkpoint is selected by micro-averaged validation accuracy.

\myparagraph{Compute.} Training uses a single NVIDIA L40S (48 GB), taking ${\sim}$9 min/epoch and ${\sim}$2.3 GPU-hours total.

\myparagraph{Convergence.} Table~\ref{tab:cnn_curve} reports per-epoch accuracy. Validation accuracy rises from 41.6\% (epoch 1) to 46.8\% (epoch 15) with train--validation gap below 1\,pp, indicating capacity bottleneck rather than overfitting. Training was capped at 15 epochs by compute budget; further epochs would yield modest gains within the same regime but are unlikely to close the gap to the 49.4\% text-only floor given the small backbone.

\begin{table}[h]
\caption{Per-epoch accuracy of the multi-task 3D CNN. Combined = micro-averaged across YN, MCQ, and Open. Bold = best validation checkpoint, used in Table~\ref{tab:baselines}.}
\label{tab:cnn_curve}
\centering
\scriptsize
\setlength{\tabcolsep}{3pt}
\begin{tabular}{rcccccccc}
\toprule
 & \multicolumn{4}{c}{Train} & \multicolumn{4}{c}{Validation} \\
\cmidrule(lr){2-5} \cmidrule(lr){6-9}
Epoch & Loss & YN & MCQ & Open & YN & MCQ & Open & Comb. \\
\midrule
1  & 1.211 & 49.8 & 24.8 & 38.3 & 50.8 & 25.9 & 46.8 & 41.6 \\
5  & 1.010 & 51.9 & 26.0 & 50.5 & 50.2 & 24.9 & 55.0 & 42.6 \\
10 & 0.986 & 55.0 & 26.2 & 53.4 & 53.7 & 25.5 & 58.5 & 45.1 \\
\textbf{15} & \textbf{0.964} & \textbf{57.6} & \textbf{26.6} & \textbf{57.2} & \textbf{56.7} & \textbf{27.0} & \textbf{57.1} & \textbf{46.8} \\
\midrule
\multicolumn{1}{l}{\textit{Test}} & --- & --- & --- & --- & 56.8 & 25.5 & 59.6 & 46.8 \\
\bottomrule
\end{tabular}
\end{table}

\myparagraph{Discussion.} Closed accuracy (43.7\%) lies below both the text-only floor (49.4\%) and clinician mean (48.9\%), and MCQ accuracy (25.5\%) equals chance. Open EM (59.6\%) is high only because the closed-vocabulary head fits the training-answer distribution. \projectname{} is not trivially solvable by a small supervised volumetric model.

\section{Complete Template Catalog}
\label{app:templates}

All 203 unique question templates organized by category. N represents subject-specific numeric values.

\begin{table}[h]
\centering
\scriptsize
\caption{Anatomy (37 templates, 6,884 items).}
\label{tab:tmpl_anatomy}
\begin{tabular}{clr}
\toprule
\# & Template & Items \\
\midrule
1 & [MCQ] Which subcortical structure in this scan shows the most volume deviation from expe... & 1,463 \\
2 & [Open] Are the ventricles relatively enlarged or within expected range in this scan (subj... & 736 \\
3 & [Open] Are the ventricles relatively enlarged or normal in this pediatric scan? (subject ... & 434 \\
4 & [MCQ] In which quartile does the total hippocampal volume fall? & 356 \\
5 & [YN] Is the right hippocampal volume above the population median (N mL)? & 338 \\
6 & [YN] Is the left hippocampal volume above the population median (N mL)? & 335 \\
7 & [YN] Is the total hippocampal volume above the population median (N mL)? & 332 \\
8 & [MCQ] Which subcortical structure shows the most volume reduction in this scan? & 302 \\
9 & [Open] Are the ventricles enlarged or normal in this scan? & 239 \\
10 & [YN] Is the entorhinal cortex thinner than expected for this N-year-old subject? & 213 \\
11 & [YN] Is the entorhinal cortical thickness within expected range for this N-year-old sub... & 212 \\
12 & [YN] Is the hippocampal volume within the expected range for this N-year-old subject? & 210 \\
13 & [YN] Is the hippocampal volume below the expected range for this N-year-old subject? & 210 \\
14 & [Open] Are the ventricles relatively enlarged or normal in this scan? (subject age: N) & 201 \\
15 & [YN] Is hippocampal atrophy present in this scan? & 169 \\
16 & [MCQ] What best describes the hippocampal morphology in this scan? & 122 \\
17 & [YN] Does this scan contain structural findings that suggest findings consistent with m... & 122 \\
18 & [YN] Are the ventricles within expected size for this N-year-old subject? & 100 \\
19 & [YN] Are the ventricles relatively enlarged for a pediatric brain? (subject age: N) & 95 \\
20 & [YN] Are the ventricles within normal size for a developing brain (N-N years old)? & 88 \\
21 & [YN] Are the ventricles relatively enlarged beyond expected age-related changes (subjec... & 66 \\
22 & [YN] Are the ventricles relatively enlarged for this age group (N-N years old)? & 54 \\
23 & [YN] Is the overall brain morphology within normal limits in this scan? & 47 \\
24 & [YN] Is the hippocampal volume below normal in this scan? & 41 \\
25 & [YN] Is the hippocampal volume below the expected range for this N-N-year-old subject? & 38 \\
26 & [YN] Is the entorhinal cortex thinner than normal in this scan? & 34 \\
27 & [YN] Is the entorhinal cortical thickness within expected range for this N-N-year-old s... & 34 \\
28 & [YN] Is the entorhinal cortex thinner than expected for this N-N-year-old subject? & 31 \\
29 & [YN] Is the hippocampal volume within the expected range for this N-N-year-old subject? & 31 \\
30 & [YN] Are there any structural abnormalities in this scan that may suggest findings cons... & 30 \\
31 & [YN] Are the ventricles normal in size relative to brain size in this scan? & 29 \\
32 & [YN] Does this scan show structural findings consistent with normal cognitive function? & 28 \\
33 & [YN] Are the ventricles within normal size for a young adult brain (N-N years old)? & 28 \\
34 & [YN] Is the entorhinal cortical thickness within normal range in this scan? & 27 \\
35 & [YN] Is the hippocampal volume within normal range in this scan? & 27 \\
36 & [YN] Are the ventricles enlarged relative to brain size in this scan? & 22 \\
37 & [YN] Are the ventricles relatively enlarged for a young adult brain? (subject age: N) & 14 \\
\bottomrule
\end{tabular}
\end{table}

\begin{table}[h]
\centering
\scriptsize
\caption{Demographic (10 templates, 8,289 items).}
\label{tab:tmpl_demographic}
\begin{tabular}{clr}
\toprule
\# & Template & Items \\
\midrule
1 & [Open] What is the sex of this subject? & 2,227 \\
2 & [MCQ] What age range does this subject likely belong to? & 2,222 \\
3 & [YN] Is this subject female? & 1,235 \\
4 & [YN] Is this subject older than N years old? & 924 \\
5 & [MCQ] What is the estimated age range of this subject? & 867 \\
6 & [YN] Is this subject male? & 406 \\
7 & [YN] Is this subject younger than N years old? & 158 \\
8 & [YN] Is this subject older than N? & 109 \\
9 & [YN] Is this subject likely a young adult (N-N years old)? & 109 \\
10 & [YN] Is this subject's age likely to be N years old or younger? & 32 \\
\bottomrule
\end{tabular}
\end{table}

\begin{table}[h]
\centering
\scriptsize
\caption{Diagnosis (18 templates, 5,573 items).}
\label{tab:tmpl_diagnosis}
\begin{tabular}{clr}
\toprule
\# & Template & Items \\
\midrule
1 & [YN] Does this brain scan from a subject (age: N) show structural features consistent w... & 903 \\
2 & [YN] Does this brain scan from a subject (age: N) show structural features inconsistent... & 837 \\
3 & [YN] Does this brain scan from a subject (age: N) show structural features consistent w... & 835 \\
4 & [MCQ] What is the most likely clinical classification based on this brain scan? & 602 \\
5 & [YN] Is this brain scan from a child with above average cognitive performance? & 499 \\
6 & [YN] Does this brain scan from a pediatric subject (age: N) show structural features co... & 496 \\
7 & [YN] Does this brain scan from a pediatric subject (age: N) show structural features co... & 495 \\
8 & [MCQ] What type of condition is most consistent with the findings in this scan? & 348 \\
9 & [YN] Does this brain scan show structural findings inconsistent with a dementia diagnos... & 126 \\
10 & [YN] Are there any structural abnormalities in this brain scan that suggest conclusive ... & 123 \\
11 & [YN] Does this brain scan show structural changes consistent with dementia? & 98 \\
12 & [YN] Does this brain scan show structural findings inconsistent with an Alzheimer's dis... & 68 \\
13 & [YN] Does this brain scan show structural abnormalities consistent with Alzheimer's dis... & 43 \\
14 & [MCQ] What is the most likely diagnosis? & 41 \\
15 & [YN] Are there any structural findings that suggest an alternative diagnosis to idiopat... & 28 \\
16 & [YN] Does this brain scan show structural findings consistent with Parkinson's disease? & 27 \\
17 & [YN] Does this brain scan show any structural abnormalities? & 27 \\
18 & [YN] Is this a brain scan from a neurologically healthy subject? & 13 \\
\bottomrule
\end{tabular}
\end{table}

\begin{table}[h]
\centering
\scriptsize
\caption{Location (28 templates, 4,526 items).}
\label{tab:tmpl_location}
\begin{tabular}{clr}
\toprule
\# & Template & Items \\
\midrule
1 & [Open] Which side has a smaller lateral ventricle volume in this brain scan? & 656 \\
2 & [Open] Which side has a smaller accumbens volume in this brain scan? & 447 \\
3 & [MCQ] In which lobe is the primary lesion located? & 409 \\
4 & [Open] Describe the location of the lesion. & 409 \\
5 & [YN] Is the right lateral ventricle smaller than the left lateral ventricle in this scan? & 310 \\
6 & [YN] Is the left lateral ventricle smaller than the right lateral ventricle in this scan? & 309 \\
7 & [MCQ] Which hippocampus appears larger in this scan? & 231 \\
8 & [Open] Which side has a smaller pallidum volume in this brain scan? & 220 \\
9 & [Open] Which side has a smaller amygdala volume in this brain scan? & 202 \\
10 & [Open] Which side has a smaller putamen volume in this brain scan? & 202 \\
11 & [YN] Is the right accumbens smaller than the left accumbens in this scan? & 183 \\
12 & [Open] Which side has a smaller hippocampus volume in this brain scan? & 180 \\
13 & [YN] Is the left accumbens smaller than the right accumbens in this scan? & 161 \\
14 & [YN] Is the lesion located in the temporal lobe? & 117 \\
15 & [Open] Which side has a smaller thalamus volume in this brain scan? & 82 \\
16 & [YN] Is the lesion located in the frontal lobe? & 82 \\
17 & [Open] Which side has a smaller caudate volume in this brain scan? & 62 \\
18 & [YN] Is the lesion located in the parietal lobe? & 62 \\
19 & [YN] Is the left pallidum smaller than the right pallidum in this scan? & 36 \\
20 & [YN] Is the right amygdala smaller than the left amygdala in this scan? & 34 \\
21 & [YN] Is the left amygdala smaller than the right amygdala in this scan? & 34 \\
22 & [YN] Is the left putamen smaller than the right putamen in this scan? & 24 \\
23 & [YN] Is the left hippocampus smaller than the right hippocampus in this scan? & 20 \\
24 & [YN] Is the right pallidum smaller than the left pallidum in this scan? & 16 \\
25 & [YN] Is the right putamen smaller than the left putamen in this scan? & 14 \\
26 & [YN] Is the right hippocampus larger than the left hippocampus in this scan? & 10 \\
27 & [YN] Is the right hippocampus smaller than the left hippocampus in this scan? & 8 \\
28 & [YN] Is the right caudate smaller than the left caudate in this scan? & 6 \\
\bottomrule
\end{tabular}
\end{table}

\begin{table}[h]
\centering
\scriptsize
\caption{Longitudinal (14 templates, 1,166 items).}
\label{tab:tmpl_longitudinal}
\begin{tabular}{clr}
\toprule
\# & Template & Items \\
\midrule
1 & [YN] Has the left hippocampal volume decreased compared to the scan N months prior (sub... & 258 \\
2 & [YN] Has the left hippocampal volume increased compared to the scan N months prior (sub... & 247 \\
3 & [YN] Has the whole brain volume increased compared to the scan N months prior (subject ... & 220 \\
4 & [YN] Has the whole brain volume decreased compared to the scan N months prior (subject ... & 206 \\
5 & [Open] What is the clinical progression status of this subject's brain compared to the pr... & 47 \\
6 & [MCQ] Has the clinical status progressed, remained stable, or improved compared to the v... & 44 \\
7 & [Open] What is the clinical progression status of this subject's brain compared to the vi... & 42 \\
8 & [MCQ] Has the subject's clinical status progressed, remained stable, or improved compare... & 40 \\
9 & [YN] Have the ventricles enlarged compared to the scan N month prior (subject age: N)? & 20 \\
10 & [YN] Have the ventricles decreased compared to the scan N months prior (subject age: N)? & 15 \\
11 & [YN] Do structural findings in this scan suggest that the subject's motor function wors... & 13 \\
12 & [YN] Do structural findings in this scan suggest that the subject's motor function impr... & 8 \\
13 & [Open] What is the cognitive trajectory of the subject's brain compared to the visit N mo... & 3 \\
14 & [MCQ] Has the subject's cognitive performance improved, remained stable, or declined com... & 3 \\
\bottomrule
\end{tabular}
\end{table}

\begin{table}[h]
\centering
\scriptsize
\caption{Measurement (11 templates, 4,771 items).}
\label{tab:tmpl_measurement}
\begin{tabular}{clr}
\toprule
\# & Template & Items \\
\midrule
1 & [MCQ] What range best describes the intracranial volume of this scan? & 2,309 \\
2 & [YN] Is the ventricular-to-brain volume ratio greater than N\% in this scan? & 396 \\
3 & [YN] Is the ventricular-to-brain volume ratio less than N\% in this scan? & 375 \\
4 & [YN] Is the average hippocampal volume below N mm³ in this scan? & 325 \\
5 & [YN] Is the average hippocampal volume above N mm³ in this scan? & 324 \\
6 & [YN] Is the total brain volume less than N mm³ in this scan? & 291 \\
7 & [YN] Is the total brain volume greater than N mm³ in this scan? & 277 \\
8 & [MCQ] What range best describes the average hippocampal volume (mm³)? & 174 \\
9 & [YN] Is the average hippocampal volume above N mm³ in this scan? & 138 \\
10 & [YN] Is the average hippocampal volume below N mm³ in this scan? & 122 \\
11 & [MCQ] What range best describes the total ventricular volume (mm³)? & 40 \\
\bottomrule
\end{tabular}
\end{table}

\begin{table}[h]
\centering
\scriptsize
\caption{Modality (2 templates, 2,621 items).}
\label{tab:tmpl_modality}
\begin{tabular}{clr}
\toprule
\# & Template & Items \\
\midrule
1 & [YN] Is this a TN-weighted MRI scan? & 2,044 \\
2 & [MCQ] What type of MRI sequence is shown in this image? & 577 \\
\bottomrule
\end{tabular}
\end{table}

\begin{table}[h]
\centering
\scriptsize
\caption{Presence (27 templates, 6,403 items).}
\label{tab:tmpl_presence}
\begin{tabular}{clr}
\toprule
\# & Template & Items \\
\midrule
1 & [MCQ] Which of the following findings is NOT present in this scan? & 1,483 \\
2 & [YN] Is ventricular enlargement present in this scan? & 456 \\
3 & [YN] Is hippocampal volume below expected range in this scan (subject age: N)? & 452 \\
4 & [YN] Are both hippocampi within expected volume range for this N-year-old subject? & 378 \\
5 & [MCQ] Which of the following findings is present in this scan? & 360 \\
6 & [YN] Is there compression of adjacent structures? & 300 \\
7 & [YN] Are the ventricles normal-sized in this scan? & 289 \\
8 & [MCQ] Which best describes the structural findings in this scan? & 276 \\
9 & [MCQ] What is the pattern of contrast enhancement? & 230 \\
10 & [YN] Is there evidence of hemorrhage within the lesion? & 224 \\
11 & [YN] Does this subject's brain show any structural deviations from age-related changes ... & 213 \\
12 & [YN] Is there evidence of perilesional edema? & 200 \\
13 & [Open] Based on this scan, does the subject have normal cognition, mild cognitive impairm... & 185 \\
14 & [YN] Is ventricular enlargement beyond expected for a N-year-old range present in this ... & 167 \\
15 & [YN] Are the ventricles within expected size in this scan (subject age: N)? & 155 \\
16 & [YN] Is hippocampal volume loss present in this scan? & 132 \\
17 & [YN] Are both hippocampi within normal volume range in this scan? & 125 \\
18 & [YN] Are the ventricles in this scan within expected size for an N-year-old subject? & 122 \\
19 & [YN] Does the lesion have well-defined borders? & 116 \\
20 & [MCQ] Which of the following findings is NOT present in this subject? & 106 \\
21 & [MCQ] Which of the following findings is present in this subject? & 82 \\
22 & [YN] Does this scan show relatively small hippocampal volumes (below Nth percentile)? & 81 \\
23 & [YN] Is this subject free of Alzheimer's disease? & 62 \\
24 & [YN] Are there any structural changes consistent with mild cognitive impairment (MCI) i... & 58 \\
25 & [YN] Is there notable hippocampal asymmetry (>N\% difference between left and right)? & 58 \\
26 & [YN] Is hippocampal volume below expected range in this scan (subject age: N-N)? & 53 \\
27 & [YN] Are both hippocampi within expected volume range for this N-N-year-old subject? & 40 \\
\bottomrule
\end{tabular}
\end{table}

\begin{table}[h]
\centering
\scriptsize
\caption{Reasoning (25 templates, 6,977 items).}
\label{tab:tmpl_reasoning}
\begin{tabular}{clr}
\toprule
\# & Template & Items \\
\midrule
1 & [Open] What is the most prominent structural finding in this brain scan? & 2,068 \\
2 & [MCQ] Which description best matches the structural findings in this scan? & 1,584 \\
3 & [MCQ] Which structural pattern best describes the findings in this scan? & 477 \\
4 & [MCQ] Which pattern best describes the relationship between this subject's age and hippo... & 356 \\
5 & [Open] What imaging features support the subject's likely diagnosis? & 351 \\
6 & [YN] Is the hippocampal volume consistent with what would be expected for this subject'... & 302 \\
7 & [MCQ] Which imaging finding most strongly supports the diagnosis of meningioma? & 228 \\
8 & [YN] Is the overall brain morphology within expected limits for this N-year-old subject? & 198 \\
9 & [MCQ] Which combination of clinical findings and diagnosis best describes this subject? & 189 \\
10 & [Open] What is the most prominent clinical finding for this subject? & 165 \\
11 & [YN] Is the overall brain morphology within expected limits for this N-N-year-old subject? & 151 \\
12 & [YN] Are there any structural findings that deviate from what is expected for a subject... & 147 \\
13 & [YN] Does the brain show structural deviations beyond expected age-related changes (sub... & 132 \\
14 & [YN] Does the brain show structural deviations from expected development within the age... & 80 \\
15 & [YN] Does the brain show structural deviations from expected developmental norms (subje... & 78 \\
16 & [YN] Is the overall brain morphology within expected limits for this N-year-old develop... & 74 \\
17 & [YN] Are the structural findings consistent with normal development for a N year-old su... & 63 \\
18 & [YN] Are there any structural abnormalities that suggest atypical development for a N-y... & 56 \\
19 & [YN] Does the brain morphology suggest that the overall cognitive function is within no... & 52 \\
20 & [YN] Do the structural patterns in this scan suggest findings associated with condition... & 52 \\
21 & [YN] Are there any structural patterns consistent with findings that may suggest mild c... & 44 \\
22 & [YN] Do the structural patterns in this scan suggest motor-related changes in the subje... & 42 \\
23 & [YN] Does the brain structure in this scan suggest neurodegenerative changes? & 35 \\
24 & [YN] Do the structural patterns in this scan suggest preserved motor function in the su... & 27 \\
25 & [YN] Do the structural findings suggest neurodegenerative changes in this subject? & 16 \\
\bottomrule
\end{tabular}
\end{table}

\section{Vision-Language Model Results}
\label{app:vlm}

Tables~\ref{tab:per_qtype}--\ref{tab:open_metrics} report per-format, per-category, and open-ended metrics for all evaluated VLMs on the non-longitudinal subset of test-public. Results are shown for text-only (no image) and the best vision condition (triplanar or stack, whichever achieves higher closed accuracy).

\begin{table}[h]
\centering
\scriptsize
\caption{Severity (19 templates, 5,580 items).}
\label{tab:tmpl_severity}
\begin{tabular}{clr}
\toprule
\# & Template & Items \\
\midrule
1 & [Open] Based on this scan, describe the expected fluid cognition level for this subject. & 1,397 \\
2 & [MCQ] How would you classify the severity of mass effect in this scan? & 458 \\
3 & [Open] Based on this scan, describe the expected cognitive screening performance of this ... & 396 \\
4 & [YN] Does this subject's scan show structural findings consistent with cognitive screen... & 383 \\
5 & [Open] What cognitive severity level is most consistent with this scan? & 354 \\
6 & [YN] Does the subject's brain scan show structural findings associated with below-avera... & 309 \\
7 & [YN] Are there any structural findings in this scan consistent with below-average globa... & 291 \\
8 & [MCQ] What disease severity level is most consistent with this brain scan? & 252 \\
9 & [YN] Does this brain scan contain structural patterns suggestive of high-grade pathology? & 224 \\
10 & [Open] What cognitive severity level is most consistent with this subject? & 217 \\
11 & [MCQ] What is the motor severity level most consistent with this brain scan? & 203 \\
12 & [MCQ] What disease severity level is most consistent with this subject? & 191 \\
13 & [YN] Are there structural patterns in the brain that suggest the fluid cognition score ... & 181 \\
14 & [YN] Does the subject's brain morphology suggest findings consistent with superior glob... & 178 \\
15 & [YN] Are there any structural findings in the scan that would be inconsistent with inta... & 149 \\
16 & [YN] Are there any structural findings in this scan that suggest reduced fluid cognitio... & 118 \\
17 & [YN] Do the structural findings in this scan suggest that the subject's cognitive impai... & 105 \\
18 & [MCQ] What stage of parkinsonism is most consistent with this subject's brain scan? & 96 \\
19 & [YN] Is the perilesional edema extensive? & 78 \\
\bottomrule
\end{tabular}
\end{table}

\begin{table}[h]
\centering
\scriptsize
\caption{Signal (12 templates, 4,163 items).}
\label{tab:tmpl_signal}
\begin{tabular}{clr}
\toprule
\# & Template & Items \\
\midrule
1 & [MCQ] What is the severity of white matter changes in this scan? & 1,135 \\
2 & [MCQ] What is the signal intensity of the lesion on TN-weighted imaging? & 688 \\
3 & [YN] Are white matter signal abnormalities present in this scan? & 642 \\
4 & [Open] Describe the signal characteristics of the lesion on the TN-weighted imaging scan. & 458 \\
5 & [YN] Is the white matter signal normal in this scan? & 386 \\
6 & [YN] Does this scan show white matter signal abnormalities? & 385 \\
7 & [YN] Does the lesion show high signal intensity on the TN-weighted imaging scan? & 256 \\
8 & [YN] Are white matter hyperintensities present in this scan? & 74 \\
9 & [MCQ] What is the TN signal intensity of the white matter lesions? & 51 \\
10 & [YN] Do the white matter lesions appear hypointense on the TN-weighted imaging scan? & 40 \\
11 & [YN] Does this scan show white matter hyperintensities? & 28 \\
12 & [YN] Is the white matter signal homogeneous (no hyperintensities) in this scan? & 20 \\
\bottomrule
\end{tabular}
\end{table}

\begin{table}[!htbp]
\centering
\caption{Per question-type metrics on NeuroQA test set vs.\ \textbf{no-image} (text-only). Cl\,=\,closed accuracy (YN\,+\,MCQ); EM\,=\,lenient normalized exact-match on open-ended items. All scores in \%. Bold\,=\,within 0.35\,pp of column maximum.}
\label{tab:per_qtype}
\renewcommand{\arraystretch}{0.92}
\setlength{\tabcolsep}{4pt}
\small
\begin{tabular}{@{}lrrrrrrrr@{}}
\toprule
& \multicolumn{4}{c}{\textit{No-image}} & \multicolumn{4}{c}{\textit{Stack}} \\
\cmidrule{2-5}\cmidrule{6-9}
Model & YN & MCQ & Cl & EM & YN & MCQ & Cl & EM \\
\midrule
\multicolumn{9}{l}{\textit{Supervised baseline (trained)}} \\[-2pt]
\quad 3D CNN (multi-task) & 48.3 & 24.5 & 38.4 & 41.9 & 56.8 & 25.5 & 43.7 & 59.6 \\
\addlinespace[3pt]
\multicolumn{9}{l}{\textit{GPT-5 family}} \\[-2pt]
\quad GPT-5.1    & 50.8 & 24.7 & 39.9 &  7.2 & 54.8 & 33.8 & 46.0 &  9.1 \\
\quad GPT-5.2    & 50.8 & 27.8 & 41.2 &  9.6 & 54.7 & \textbf{34.7} & 46.3 & 14.6 \\
\quad GPT-5-mini & 50.9 & 28.5 & 41.5 &  7.8 & 54.3 & 33.9 & 45.8 & 12.4 \\
\addlinespace[3pt]
\multicolumn{9}{l}{\textit{GPT-4 family}} \\[-2pt]
\quad GPT-4o       & 52.6 & 27.6 & 42.2 & 15.5 & 54.4 & 33.8 & 45.8 & 15.0 \\
\quad GPT-4.1      & 52.8 & 27.8 & 42.3 & 17.9 & 54.6 & 33.7 & 45.8 & 18.0 \\
\quad GPT-4.1-mini & 50.8 & 22.9 & 39.1 & 20.3 & 52.4 & 25.9 & 41.2 & 13.9 \\
\quad GPT-4.1-nano & 46.1 & 25.3 & 37.3 & 13.2 & 47.7 & 24.8 & 38.1 & \textbf{24.3} \\
\addlinespace[3pt]
\multicolumn{9}{l}{\textit{Gemini}} \\[-2pt]
\quad Gemini-3.1-Pro & 49.3 & 24.9 & 39.0 &  6.5 & \textbf{58.0} & 33.5 & \textbf{47.7} & 23.2 \\
\quad Gemini-2.5-Pro & 51.8 & 27.5 & 41.6 & 14.2 & 54.0 & 27.2 & 42.7 &  9.9 \\
\addlinespace[3pt]
\multicolumn{9}{l}{\textit{LLaMA-4}} \\[-2pt]
\quad LLaMA-4-Mav.  & 51.9 & 26.4 & 41.2 & \textbf{25.5} & 52.7 & 26.4 & 41.6 & 18.0 \\
\quad LLaMA-4-Scout & 51.2 & \textbf{29.9} & 42.3 & \textbf{25.3} & 53.3 & 31.1 & 44.0 & \textbf{24.4} \\
\addlinespace[3pt]
\multicolumn{9}{l}{\textit{Anthropic}} \\[-2pt]
\quad Claude Sonnet 4.6 & 48.7 & 26.7 & 39.4 &  7.5 & 51.7 & 29.4 & 42.3 & 21.8 \\
\addlinespace[3pt]
\multicolumn{9}{l}{\textit{Medical}} \\[-2pt]
\quad MedGemma-27B & 38.3 & 19.3 & 30.3 & 12.4 & 48.9 & 26.7 & 39.6 & \textbf{24.8} \\
\quad MedGemma-4B  & 47.2 & 26.7 & 38.6 & 16.1 & 41.1 & 27.8 & 35.6 & 18.6 \\
\bottomrule
\end{tabular}
\end{table}

\begin{table}[!htbp]
\centering
\caption{Per-category closed-ended accuracy (\%) on NeuroQA. Bold\,=\,within 0.35\,pp of column maximum. Category abbreviations: Dem\,=\,Demographic, Rea\,=\,Reasoning, Ana\,=\,Anatomy, Pre\,=\,Presence, Sev\,=\,Severity, Dia\,=\,Diagnosis, Mea\,=\,Measurement, Loc\,=\,Location, Sig\,=\,Signal, Mod\,=\,Modality.}
\label{tab:per_cat}
\renewcommand{\arraystretch}{0.95}
\setlength{\tabcolsep}{3pt}
\scriptsize
\begin{tabular}{lrrrrrrrrrrrr}
\toprule
Model & Dem & Rea & Ana & Pre & Sev & Dia & Mea & Loc & Sig & Mod & Long. & Whole \\
\midrule
\multicolumn{13}{l}{\textit{Supervised baseline (trained)}} \\[-2pt]
\quad 3D CNN (multi-task) & \textbf{43.4} & 38.5 & 44.0 & \textbf{50.2} & 35.4 & 48.8 & 44.3 & 39.1 & 41.7 & 43.1 & 44.6 & 43.7 \\
\addlinespace[3pt]
\multicolumn{13}{l}{\textit{GPT-5 family}} \\[-2pt]
\quad GPT-5.1    & 38.5 & 37.9 & \textbf{51.1} & 47.0 & 37.8 & 48.6 & 36.7 & 43.8 & 49.9 & 80.2 & 51.5 & 46.1 \\
\quad GPT-5.2    & 40.2 & 39.4 & 50.3 & 42.8 & 40.8 & \textbf{51.0} & 42.4 & 38.4 & 49.0 & 79.8 & 48.5 & 46.5 \\
\quad GPT-5-mini & 39.2 & 40.6 & 43.0 & 41.8 & 39.1 & 47.9 & \textbf{46.0} & 44.3 & \textbf{51.8} & 79.4 & 51.5 & 45.9 \\
\addlinespace[3pt]
\multicolumn{13}{l}{\textit{GPT-4 family}} \\[-2pt]
\quad GPT-4o       & 39.7 & 36.5 & 49.1 & 44.1 & 38.3 & 49.5 & 43.8 & 42.4 & 48.0 & 75.5 & 54.1 & 45.9 \\
\quad GPT-4.1      & 38.4 & 38.9 & 49.5 & 42.5 & 38.3 & 49.0 & 43.3 & 42.4 & 48.8 & 80.2 & 51.5 & 46.0 \\
\quad GPT-4.1-mini & 37.9 & 37.5 & 40.9 & 38.7 & 40.2 & 45.8 & 39.5 & 44.3 & 45.0 & 45.5 & 51.5 & 41.5 \\
\quad GPT-4.1-nano & 30.9 & 31.0 & 41.8 & 36.1 & 33.5 & 46.9 & 34.0 & 42.9 & 40.1 & 50.6 & 44.6 & 38.6 \\
\addlinespace[3pt]
\multicolumn{13}{l}{\textit{Gemini}} \\[-2pt]
\quad Gemini-3.1-Pro & 40.4 & \textbf{41.1} & 47.9 & 42.5 & 38.1 & \textbf{50.8} & 39.8 & \textbf{54.1} & 48.8 & \textbf{99.2} & \textbf{57.4} & \textbf{47.9} \\
\quad Gemini-2.5-Pro & 34.0 & 36.5 & 35.8 & 42.8 & 34.3 & 48.2 & 42.4 & 49.3 & 42.0 & 79.8 & 56.4 & 42.9 \\
\addlinespace[3pt]
\multicolumn{13}{l}{\textit{LLaMA-4}} \\[-2pt]
\quad LLaMA-4-Mav.  & 37.9 & 37.7 & 37.7 & 42.3 & 38.1 & 50.5 & 30.4 & 45.3 & 37.6 & 69.6 & 47.5 & 41.8 \\
\quad LLaMA-4-Scout & 40.9 & 31.0 & 50.3 & 40.2 & \textbf{41.6} & 49.5 & 40.1 & 41.9 & 47.4 & 58.5 & 56.4 & 44.1 \\
\addlinespace[3pt]
\multicolumn{13}{l}{\textit{Anthropic}} \\[-2pt]
\quad Claude Sonnet 4.6 & 40.4 & 38.4 & 44.0 & 37.5 & 38.6 & 48.4 & 33.1 & 45.8 & 43.6 & 60.5 & 55.4 & 42.5 \\
\addlinespace[3pt]
\multicolumn{13}{l}{\textit{Medical}} \\[-2pt]
\quad MedGemma-27B & 40.2 & 35.3 & 45.6 & 38.7 & 35.9 & 46.2 & 34.7 & 48.8 & 48.2 &  9.1 & \textbf{57.4} & 39.8 \\
\quad MedGemma-4B  & 30.2 & 29.4 & 36.5 & 36.4 & 37.0 & 46.2 & 34.9 & 44.3 & 35.1 & 16.6 & 54.5 & 35.6 \\
\bottomrule
\end{tabular}
\end{table}

\begin{table}[!htbp]
\centering
\caption{Open-ended question metrics on NeuroQA. EM\,=\,exact match after lenient normalization; F1\,=\,token-level F1. All scores in \%. Bold\,=\,within 0.35\,pp of column maximum. }
\label{tab:open_metrics}
\renewcommand{\arraystretch}{0.92}
\small
\begin{tabular}{@{}l@{\hskip 6pt}r@{\hskip 4pt}r@{\hskip 8pt}r@{\hskip 4pt}r@{}}
\toprule
& \multicolumn{2}{c}{\textit{No-image}} & \multicolumn{2}{c}{\textit{Stack}} \\
\cmidrule{2-3}\cmidrule{4-5}
Model & EM & F1 & EM & F1 \\
\midrule
\multicolumn{5}{l}{\textit{Supervised baseline (trained)}} \\[-2pt]
\quad 3D CNN (multi-task) & 41.9 & 51.9 & 59.6 & 67.3 \\
\addlinespace[3pt]
\multicolumn{5}{l}{\textit{GPT-5 family}} \\[-2pt]
\quad GPT-5.1    &  7.2 & 14.8 &  9.1 & 16.7 \\
\quad GPT-5.2    &  9.6 & 14.3 & 14.2 & 19.1 \\
\quad GPT-5-mini &  7.8 & 10.5 & 12.4 & 18.1 \\
\addlinespace[3pt]
\multicolumn{5}{l}{\textit{GPT-4 family}} \\[-2pt]
\quad GPT-4o       & 15.5 & 18.8 & 15.0 & 18.1 \\
\quad GPT-4.1      & 17.9 & 20.1 & 18.0 & 23.4 \\
\quad GPT-4.1-mini & 20.3 & 21.5 & 13.9 & 19.5 \\
\quad GPT-4.1-nano & 13.2 & 14.1 & \textbf{24.3} & 25.2 \\
\addlinespace[3pt]
\multicolumn{5}{l}{\textit{Gemini}} \\[-2pt]
\quad Gemini-3.1-Pro &  6.5 &  7.4 & 23.2 & 26.6 \\
\quad Gemini-2.5-Pro & 14.2 & 18.2 &  9.9 & 14.5 \\
\addlinespace[3pt]
\multicolumn{5}{l}{\textit{LLaMA-4}} \\[-2pt]
\quad LLaMA-4-Mav.  & \textbf{25.5} & 26.5 & 18.0 & 21.1 \\
\quad LLaMA-4-Scout & \textbf{25.3} & \textbf{32.8} & \textbf{24.4} & \textbf{30.0} \\
\addlinespace[3pt]
\multicolumn{5}{l}{\textit{Anthropic}} \\[-2pt]
\quad Claude Sonnet 4.6 &  7.5 &  7.7 & 21.8 & 27.6 \\
\addlinespace[3pt]
\multicolumn{5}{l}{\textit{Medical}} \\[-2pt]
\quad MedGemma-27B & 12.4 & 14.5 & \textbf{24.8} & 28.6 \\
\quad MedGemma-4B  & 16.1 & 16.2 & 18.6 & 18.8 \\
\bottomrule
\end{tabular}
\end{table}

\end{document}